\documentclass[sn-mathphys-num]{sn-jnl}

\usepackage[algo2e,ruled,vlined,linesnumbered]{algorithm2e}
\usepackage{graphicx}%
\usepackage{multirow}%
\usepackage{caption} 
\usepackage{amsmath,amssymb,amsfonts}%
\usepackage{amsthm}%
\usepackage{mathrsfs}%
\usepackage[title]{appendix}%
\usepackage{xcolor}%
\usepackage{textcomp}%
\usepackage{manyfoot}%
\usepackage{booktabs}%
\usepackage{algorithm}%
\usepackage{algorithmicx}%
\usepackage{algpseudocode}%
\usepackage{listings}%
\usepackage{xspace}

\newcommand{\ooea}{$(1 + 1)$-EA\xspace}
\newcommand{\opoea}{\ooea}
\newcommand{\olea}{$(1 + \lambda)$-EA\xspace}
\newcommand{\oplea}{\olea}
\newcommand{\oclea}{$(1 , \lambda)$-EA\xspace}
\newcommand{\moea}{$(\mu + 1)$-EA\xspace}

\newcommand{\saoplea}{\text{SA-}$(1 + \lambda)$-EA\xspace}
\newcommand{\saolea}{\text{SA-}$(1, \lambda)$-EA\xspace}

\newcommand{\fdbv}{\textsc{FDBV}\xspace}
\newcommand{\adbv}{\textsc{ADBV}\xspace}
\newcommand{\sdbv}{\textsc{SDBV}\xspace}
\newcommand{\poea}{\textsc{PO-EA}\xspace}
\newcommand{\poeam}{\textsc{PO-EA\(^-\)}\xspace}

\newcommand{\onemax}{\textsc{OneMax}\xspace}

\newcommand{\OM}{\textsc{Om}\xspace}
\newcommand{\ZM}{Z}

\newcommand{\BinVal}{\textsc{BinVal}\xspace}
\newcommand{\Binval}{\textsc{BinVal}\xspace}
\newcommand{\dynBV}{\textsc{Dynamic BinVal}\xspace}

\newcommand{\hottopic}{\textsc{HotTopic}\xspace}

\newcommand{\xinit}{\ensuremath{x^{\mathrm{init}}}}

\newcommand{\EE}{\ensuremath{\mathbf{E}}}

\newcommand{\indicator}[1]{\ensuremath{\mathbf{1}_{#1}}}

\newcommand{\cE}{\ensuremath{\mathcal{E}}}
\newcommand{\Bin}{\ensuremath{\mathrm{Bin}}}
\newcommand{\Var}{\ensuremath{\mathrm{Var}\,}}

\DeclareMathOperator*{\argmax}{arg\,max}

\newcommand{\RR}{\ensuremath{\mathbb{R}}}

\newcommand\eps{\varepsilon}

\newcommand{\proofitem}[1]{$\blacktriangleright$~\textbf{#1:}}

\theoremstyle{thmstyleone}%
\newtheorem{theorem}{Theorem}
\theoremstyle{thmstyletwo}%
\newtheorem{question}{Question}%
\newtheorem{conjecture}{Conjecture}
\newtheorem{lemma}{Lemma}

\theoremstyle{thmstylethree}%

\begin{document}

\title[Article Title]{Hardest Monotone Functions for Evolutionary Algorithms}


\author*{\fnm{Marc} \sur{Kaufmann}}\email{marc.kaufmann@inf.ethz.ch}

\author*{\fnm{Maxime} \sur{Larcher}}\email{maxime.larcher@inf.ethz.ch}

\author*{\fnm{Johannes} \sur{Lengler}}\email{johannes.lengler@inf.ethz.ch}

\author*{\fnm{Oliver} \sur{Sieberling}}\email{osieberling@student.ethz.ch}

\affil{\orgdiv{Department of Computer Science}, \orgname{ETH Z{\"u}rich}, \orgaddress{\country{Switzerland}}}


\abstract{In this paper we revisit the question how hard it can be for the $(1+1)$ Evolutionary Algorithm to optimize monotone pseudo-Boolean functions. By introducing a more pessimistic stochastic process, the partially-ordered evolutionary algorithm (\poea) model, Jansen first proved a runtime bound of $O(n^{3/2})$. More recently, Lengler, Martinsson and Steger improved this upper bound to $O(n \log^2 n)$ by an entropy compression argument. In this work, we analyze monotone functions that may adversarially vary at each step of the optimization, so-called dynamic monotone functions.
We introduce the function Switching Dynamic BinVal (SDBV) and prove, using a combinatorial argument, that for the \opoea with any mutation rate \(p \in [0,1]\), SDBV is drift minimizing within the class of dynamic monotone functions. We further show that the \ooea optimizes SDBV in $\Theta(n^{3/2})$ generations. Therefore, our construction provides the first explicit example which realizes the pessimism of the \poea model. Our simulations demonstrate matching runtimes for both the static and the self-adjusting \oclea and \oplea. Moreover, devising an example for fixed dimension, we illustrate that drift minimization does not equal maximal runtime beyond asymptotic analysis. }

\keywords{evolutionary algorithm, $(1 + 1)$-EA, dynamic environments, hardest functions, precise runtime analysis, drift minimization}



\maketitle

\section{Introduction}
Evolutionary Algorithms are randomized optimization algorithms. They originate in practical applications, and since they are often employed on NP-hard problems, there can be no general theory proving their effectiveness. However, for the understanding of such algorithms, it has turned out fruitful to analyze them theoretically on benchmark functions. The benchmark function should be simple enough that some randomized algorithms find the optimum efficiently, because then it is possible to study whether other algorithms show a similar performance. A specific, intensively studied class of such functions are pseudo-Boolean \emph{monotone} functions. In such functions, replacing a zero-bit by a one-bit improves the objective value of any search point. Thus they are easy enough to be optimized efficiently by \emph{some} algorithms. However, it has turned out that many algorithms used in practice fail to optimize this class of functions efficiently~\cite{doerr2010optimizing,lengler2018drift,lengler2019general, lengler2021exponential}, hence this set of benchmark functions has high discriminative power. 

The first analysis of an algorithm on the family of all monotone functions goes back to Jansen~\cite{jansen2007brittleness}. In order to study the \opoea --- one of the simplest evolutionary algorithms that maintains only a single individual as population, see Section~\ref{sec:algorithms} for details --- Jansen introduced a random process which he called the Partially-Ordered Evolutionary Algorithm (\poea). This process is a pessimistic model for how the \opoea behaves on any monotone function, in particular for the sub-class of linear functions that was the focus of~\cite{jansen2007brittleness}. The process was named after its key component, a partial ordering on the search space $\{0,1\}^n$. 
The \poea "shares" with all monotone functions the unique global optimum, in the sense that the string consisting of only 1s is the unique search point which the PO-EA will not replace by any offspring. However, it does not operate on a fitness landscape defined by any fitness function.


One potential drawback of this model is that it may be overly pessimistic compared to the most difficult fitness landscapes which the $(1+1)$-EA may encounter. To see why, let us first explain informally how the \poea operates. 

We say that a search point $x \in \{0,1\}^n$ \textit{dominates} another search point $y\in \{0,1\}^n$ if it dominates $y$ bit-wise, that is, there is no position $i$ such that $y_i=1$ but $x_i=0$. Many search points are not comparable in this way --- for instance with $n=3$, $x=101$ and $y=010$ are not comparable --- so this indeed only defines a partial order on the search space. Still, the string consisting of all 1s is comparable to and dominates any other search point, which motivates the choice of this partial order as a measure of optimization progress. 
The \poea works as follows: in any situation where the parent and the offspring are \emph{comparable}\footnote{That is, either the offspring dominates the parent, or vice-versa.}, the parent for the subsequent generation is chosen as the one that dominates the other --- note that in this case, the selection indeed mimics that of the optimization by the \opoea of any monotone pseudo-Boolean function. When the parent and offspring are \emph{incomparable}, then the \poea chooses as the next parent the one with fewer \(1\)-bits.

In other words, in this second case, the algorithm deliberately stays or moves away from the optimum in terms of Hamming distance and thus gives a pessimistic estimate of optimization progress. The PO-EA model was subsequently extended by Colin, Doerr and Férey to enable parametrizing the degree of pessimism~\cite{colin2014monotonic}. In their model, with some probability $q$ the random process takes the pessimistic route and opts for standard selection, that is, picking the survivor based on the Hamming distance to the optimum otherwise. 

The PO-EA model is useful because it gives upper bounds\footnote{To be absolutely formal, the runtime of the \opoea on any monotone function is stochastically dominated by that of the \poeam, a variant of the \poea introduced in~\cite{colin2014monotonic}. In most cases (and even more in the hardest stages of the optimization when close to the optimum) the \poea and \poeam behave in very similar ways, so we ignore this subtlety in the rest of this introduction.} for the runtime of the $(1+1)$-EA on some of the most intensively studied classes of functions: \emph{linear functions}~\cite{witt2013tight} (including the popular benchmarks \onemax and \BinVal) and \emph{monotone functions}~\cite{doerr2013mutation,lengler2019general} (including the class of \hottopic functions). It also applies to recent generalizations of these classes to dynamic environments~\cite{lengler2018noisy,lengler2021runtime,kaufmann2023self_TCS}. In particular, the results by Jansen~\cite{jansen2007brittleness} and Colin, Doerr and Férey~\cite{colin2014monotonic} imply that the \ooea with standard bit mutation finds the optimum in all of these cases in time $O(n\log n)$ if the mutation rate is $\chi/n$ for any constant $\chi < 1$, and in time $O(n^{3/2})$ if\footnote{The latter result was claimed in~\cite{jansen2007brittleness}, but the proof contained an error. It was later proven in \cite{colin2014monotonic}.} $\chi=1$. However, for $\chi>1$ it was shown both for linear and for static monotone functions that those are strictly easier than the PO-EA model. In particular, for $\chi = 1$ the \ooea can optimize linear functions in $O(n\log n)$ steps~\cite{witt2013tight} and static monotone functions in $O(n\log^2 n)$ steps~\cite{lengler2019does}, while the PO-EA needs~$\Theta(n^{3/2})$ steps~\cite{colin2014monotonic}. Moreover, the \ooea can solve both linear and (static) monotone functions efficiently for some values $\chi>1$ for which the PO-EA gives exponential upper bounds. This leads to the following question.
\begin{question}[Colin, Doerr, Férey~\cite{colin2014monotonic}]
    Is the PO-EA a too pessimistic model for any monotonic function or is there a not-yet discovered monotonic function with optimization time $\Theta(n^{3/2})$?
\end{question}

While the above question was phrased with static monotone functions in mind, the aforementioned results in~\cite{lengler2019does} show that only \emph{dynamic} monotone functions might possibly achieve this bound. 
Answering this question would allow to "extend" the definition of the \poea (or at least its \emph{concept}) to other optimization algorithms that maintain a population with more than one element, or that generate more than one child at each step. Knowing the \emph{hardest} monotone functions in terms of expected optimization time or at least a hard function for the \opoea would provide a natural `hard' candidate for other algorithms, and notably its extensions such as the \moea,  the \olea or the \oclea.
In the context of \emph{self-adjusting} optimization, Kaufmann, Larcher, Lengler and Zou~\cite{kaufmann2023self_TCS} conjectured that \textsc{Adversarial \dynBV} (\adbv for short, see Section~\ref{sec:prelim} for a formal definition) is the hardest (dynamic) monotone function to optimize. While their question was only formulated for the \saolea, it naturally extends to other optimization algorithms. 
\begin{question}[Kaufmann, Larcher, Lengler, Zou~\cite{kaufmann2023self_TCS}]
    What is the hardest monotone function to optimize for the \opoea and for other optimization algorithms?
\end{question}
In this paper we will address both these questions and show that there exist dynamic monotone functions which are asymptotically as hard as the PO-EA model. 
Regarding the second question, we show that the conjectured \adbv is \emph{not} the hardest function to optimize for the \opoea in two aspects. 

Firstly, we show that the ADBV construc\-tion is not optimal when the distance from the optimum is larger than $n/2$, and we instead propose a modification which we call Switching Dynamic BinVal (SDBV). For this modified candidate, we then prove that among all dynamic monotone functions, SDBV has the minimal drift (expected progress towards the optimum), and that it does indeed have a runtime of $\Theta(n^{3/2})$ for $\chi = 1$, matching the PO-EA. This gives an answer to Question 1. Nevertheless, we also show that SDBV is in general \emph{not} the hardest dynamic monotone function \emph{in terms of runtime}. Indeed, we show by numerical calculations that it is not the hardest function for any odd $n\in[9,45]$. Hence, we give a \emph{negative} answer to the following, plausible question.

\begin{question}
    In the class of dynamic monotone functions, is the function which minimizes the drift towards the optimum also the function which leads to the largest runtime?
\end{question}

This result is especially striking because drift analysis has become the main workhorse in the analysis of evolutionary algorithms~\cite{lengler2020drift}. Our negative answer to Question 3 can be seen as complementary to the work of Buskulic and Doerr~\cite{buskulic2021maximizing}, who asked a similar question for \emph{algorithms} instead of \emph{landscapes}. More precisely, they showed that for the \ooea on \onemax, for most fitness levels between $n/2$ and $2n/3$ the optimal mutation strengths are higher than those which maximize drift. This disproved a conjecture made in \cite{doerr2018simple}. Hence, the optimal algorithm is more risk-affine than drift-maximizing one. The result in~\cite{buskulic2021maximizing} was driven by the desire for \emph{precise runtime analysis} beyond the leading order asymptotics, in particular the result by Doerr, Doerr and Yang that the unary unbiased blackbox complexity of \onemax is $n \ln n-cn+/-o(n)$ for a constant $0.2539\le c \le 0.2665$ \cite{doerr2020optimal}. As \cite{buskulic2021maximizing} show that the best \emph{algorithm} is not the drift-maximizing one, we show that the hardest \emph{fitness landscape} is not the drift-minimizing one.

\subsection{Summary of our Contribution}
The following summarizes our contribution:
\begin{itemize}
    \item We provide an explicit construction of a function called \textsc{Switching Dynamic BinVal} (\sdbv) and show that it minimizes the drift in the number of zeros at every point; we also provide a runtime analysis. This is the first construction of an explicit function which matches the bounds derived by the PO-EA framework with parametrized pessimism. In particular, \sdbv is the first dynamic monotone function which cannot be optimized by the considered class of algorithms in $\Tilde{O}(n)$ generations\footnote{Here the $\Tilde{O}(n)$ hides potential polylogarithmic factors of $n$}.
    \item We give a proof of its drift minimization - in its function class - using exclusively combinatorial arguments
    
     \item We show that drift minimization does not imply maximization of expected runtime, thus extending our understanding of hardness of fitness landscapes.

    \item We complement our findings with simulations which match our proved bounds, extending experimentally also the results to a larger class of algorithms: (1+1)-EA, ($1+\lambda$)-EA, ($1,\lambda$)-EA, \saolea, see Section~\ref{sec:algorithms} for their definitions. This provides evidence towards a verification of a wider conjecture made by Kaufmann, Larcher, Lengler and Zou in~\cite{kaufmann2023self_TCS}. 
       
\end{itemize}
We begin by stating our first main result, which holds for any mutation rate.

\begin{theorem}
    \label{thm:sdbv-minimizes-drift}
    \textsc{Switching Dynamic Binary Value} is the dynamic monotone function which minimizes the drift in the number of zeros at every search point for the \opoea with any mutation rate.
\end{theorem}

Next, we show that, perhaps surprisingly, the minimization of drift does not equate to maximal expected runtime:
\begin{theorem}
    \label{thm:drift-minimization-time-minimization}
    The drift-minimizing function for the \opoea is not the function with the largest expected optimization time.
  
\end{theorem}

The computation of the expected optimization time of Switching Dynamic Binary Value is our final key result.
\begin{theorem}
    \label{thm:sdbv-runtime}
    The \opoea with standard mutation rate $1/n$ optimizes \textsc{Switching Dynamic Binary Value} in expected $\Theta(n^{3/2})$ generations.
\end{theorem}

\subsection{Related Work}\label{sec:relatedwork}
\paragraph{Easiest and Hardest Functions.} 
In~\cite{doerr2012multiplicative}, Doerr, Johannsen and Winzen showed that \onemax is the easiest among all functions with unique global optimum on $\{0,1\}^n$ for the \ooea, meaning that it minimizes the runtime. This has been extended to other algorithms~\cite{sudholt2012new} and has been strengthened into stochastic domination formulations in various situations \cite{witt2013tight,corus2017easiest,jorritsma2023comma}. On the other hand, it has recently been shown that \onemax is not the easiest function with respect to the probability of increasing the fitness, which is important for self-adapting algorithms that rely on this probability~\cite{kaufmann2022onemax}.

While the hardest among \emph{all} pseudo-Boolean functions is also known~\cite{he2014easiest}, it is a long-standing open problem to find the hardest \emph{linear} function. A long-standing folklore conjecture is that this is the \BinVal function, but as of today, this conjecture remains unresolved.

\paragraph{Linear Functions.} The seminal paper of Witt~\cite{witt2013tight} showed that all linear functions are optimized in time~$O(n \log n)$ by the \ooea with standard bit mutation of rate $\chi/n$, for any constant $\chi$. Recently, Doerr, Janett and Lengler have generalized this to arbitrary unbiased mutation operators as long as the probability of flipping exactly one bit is $\Theta(1)$, and the expected number of flipped bits is constant~\cite{doerr2023tight}. It remains open whether this result extends to so-called \emph{fast mutation operators}~\cite{doerr2017fast} if they have unbounded expectation. Nevertheless, it is well understood that this class of functions is considerably easier than the PO-EA framework.

\paragraph{Monotone Functions.}
Monotone functions are a natural generalization of linear functions and one of the few general classes of fitness landscapes that have been thoroughly theoretically investigated. The first result was by Doerr, Jansen, Sudholt, Winzen and Zarges~\cite{doerr2013mutation}, who showed that for a mutation rate of $\chi/n$ with $\chi \ge 13$, the \ooea needs exponential time to find the optimum. The constant $13$ was later improved to $2.13..$~\cite{lengler2018drift}, and the result was generalized to a large class of other algorithms by Lengler~\cite{lengler2019general}. He invented a special type of monotone functions called \hottopic functions, which proved to be a hard instance for many evolutionary algorithms if the mutation rate is large. Moreover, Lengler and Zou showed that \hottopic instances are also hard for arbitrarily \emph{small} constants $\chi >0$ for the \moea if it is used with a large constant population size~\cite{lengler2021exponential}. As discussed earlier, all upper bounds from the PO-EA framework apply for the \ooea on all monotone functions, including the $O(n^{3/2})$ upper bound for $\chi = 1$. However, Lengler, Martinsson and Steger could strengthen this bound to $O(n\log^2 n)$, showing that (static) monotone functions are strictly easier than the PO-EA framework. Moreover, they showed that the bound $O(n\log^2 n)$ also holds for $\chi = 1+\eps$ for a sufficiently small constant $\eps >0$. In this regime, the PO-EA only gives exponential upper bounds. 

\paragraph{Dynamic Monotone Functions.} Dynamic Monotone Functions were first considered by Lengler and Schaller~\cite{lengler2018noisy}, who studied linear functions with positive weights, where the weights were re-drawn in each generation. It was later argued that the Dynamic Binary Value Function \dynBV could be obtained as a limit from dynamic linear functions if the weights are maximally skewed~\cite{lengler2020large,lengler2021runtime}. Those instances can also be hard to optimize for the \ooea if the mutation rate is large. This is even true if the dynamic environment switches always between the same two functions~\cite{janett2023two}. Dynamic Monotone Functions were then used in~\cite{kaufmann2023self_TCS} to provide instances on which a self-adapting $(1,\lambda)$-EA may fail. Due to those hardness results, dynamic functions are natural candidates for hard functions within the PO-EA framework. Indeed,~\cite{kaufmann2023self_TCS} also introduced the ADBV construction and conjectured that it is hardest among all dynamic monotone functions. We disprove this conjecture in this paper, but prove that a variant, \sdbv, is drift-minimizing, though not hardest. 

\subsection{Organization}
The remainder of the paper is organized as follows: we begin with some basic  definitions and the description of algorithms as well as the benchmark functions in Section 2. In Section 3 we construct the function Switching Dynamic BinVal and show that it minimizes Hamming drift in the class of dynamic monotone functions. Section 4 follows with an explicit computation which shows that despite minimizing the drift, SDBV does not maximize the expected runtime in its function class. In Section 5, we then show matching upper and lower bounds for its runtime. Section 6 contains our experiments, followed by the Conclusion in Section 7.

An extended abstract with the results of this paper was published at the 24th European Conference on Evolutionary Computation in Combinatorial Optimization, EvoCOP 2024, held as part of EvoStar 2024 \cite{kaufmann2024hardest}. This contained about 50\% of the material presented here. In particular, it contained the main results, but most proofs were only outlined and some experimental results were omitted. 

\section{Preliminary}\label{sec:prelim}

Throughout this paper, \(n\) denotes an arbitrary positive integer, and we consider a stochastic process on the search space \(\{0,1\}^n\). 
Given \(x \in \{0,1\}^n\), we let \(Z(x) = n - \sum_{i=1}^n x_i\) as the number of zero-bits of \(x\). For \(x, y \in \{0,1\}^n\), we say that \(x\) \emph{dominates} \(y\) --- which we write \(x \ge y\) --- if \(x_i \ge y_i\) for all \(1 \le i \le n\). A function is called \emph{monotone} if \(f(x) > f(y) \) whenever \(x\) dominates \(y\), with strict inequality in at least one of the components. A sequence of functions \((f_t)_{t \ge 0}\) is said to be \emph{dynamic monotone} if \(f_t\) is monotone for each \(t\). We abuse the terminology slightly and talk of \emph{a dynamic monotone function} rather than a dynamic monotone sequence of functions.

\subsection{Algorithms}\label{sec:algorithms}
The theoretical analysis presented in this paper focuses on the classical \opoea with standard bit mutation and mutation rate $1/n$, as follows. At each step, the parent \(x\) generates a single child \(y\) by flipping every bit with probability \(1/n\). If the fitness improves, i.e.\ if \(f(y) > f(x)\), then \(y\) is selected as the new parent for the next generation. When the fitness stagnates, i.e.\ if \(f(y) = f(x)\), we also choose $y$ as the new offspring. Otherwise \(y\) is discarded and \(x\) remains the parent for the next generation. The pseudocode for the \opoea may be found in Algorithm~\ref{alg:opoea}.
\begin{algorithm}
    \caption{\opoea with mutation rate \(p\), initial start point \(\xinit\) for maximizing a fitness function \(f \colon \{0,1\}^n \to \RR\). \label{alg:opoea}}
    \SetKwInput{Init}{Initialization}
    \SetKwInput{Mut}{Mutation}
    \SetKwInput{Sel}{Selection}
    \SetKwInput{Opt}{Optimization}
    \Init{Set $x^0= \xinit$;}
    \Opt{
    	\For{$t = 0,1,\dots$}{
    		\Mut{
                \(y^t \leftarrow \text{mutate } x^t\) by flipping each bit indep.\ with probability \(p\).
            }
            \Sel{
            \If{$f(y^t)\ge f(x^t)$}{$x^{t+1} \leftarrow y^t$}
            \Else{$x^{t+1} \leftarrow x^t$}
            }
    	}
    }
\end{algorithm}
While our proofs are restricted to the \opoea, our simulations described in Section~\ref{sec:simulations} confirm our theoretical findings for a larger class of algorithms. Namely we provide simulations for the \oplea, the \oclea, as well as their self-adjusting versions. 

The \oplea works similarly to the \opoea except that at each step~\(t\) we generate \(\lambda\) children \(y^{1,t}, \dots, y^{\lambda, t}\) instead of a single one (still by mutating every bit independently with probability \(1/n\)). As the next parent, we choose \(x^{t+1} = \argmax \{f(x^t), f(y^{1,t}), \dots, f(y^{\lambda, t})\}\), i.e.\ we choose among the offspring and the parent the one that maximizes the fitness. 
The \oclea is the \emph{non-elitist} version of the \oplea: the generation is the same but the selection is made only among the children, i.e.\ \(x^{t+1} = \argmax\{f(y^{1,t}), \dots, f(y^{\lambda, t})\}\). 
Those algorithms are now classical, and their runtime on monotone functions is well understood. We refer the interested reader to \cite{jagerskupper2007plus, neumann2009theoretical, doerr2013how, jansen2005choice, lassig2010general} and \cite{jagerskupper2007plus, neumann2009theoretical, rowe2014choice, doerr2018runtime,lengler2019general} respectively for the exact pseudocode and known results on those algorithms.

The Self-Adjusting \oplea and Self-Adjusting \oclea{} --- abbreviated \saoplea and \saolea respectively --- work by updating the size of the offspring \(\lambda^t\) at each step. Two parameters \(F>1\) and \(s>0\) control how fast \(\lambda\) changes: in case of a successful fitness improvement, the offspring population size for the following step generation decreases to \(\lambda^{t+1} \leftarrow \lambda^t / F\), and otherwise it increases to \(\lambda^{t+1} \leftarrow F^{1/s} \lambda^{t}\). Those algorithms have recently been under scrutiny and we refer to~\cite{kaufmann2022self_PPSN, kaufmann2023self_TCS, kaufmann2022onemax, hevia2021self, lassig2011adaptive, doerr20171+} for the pseudocodes and known results.

\subsection{Benchmarks}
\label{sec:benchmarks}

Monotone pseudo-Boolean functions serve as a classical benchmark to understand how evolutionary algorithms behave and how fast optimization happens. This choice is motivated by the fact that it is a large class of functions, for which theoretical analysis is `relatively easy' (or at least possible). Despite their seeming simplicity (flipping a \(0\)-bit to \(1\) always increases the fitness), the optimization is non-trivial, as was shown in a series of papers~\cite{doerr2010optimizing, doerr2013mutation,lengler2019does,lengler2019general,lengler2021exponential}. 
Two notable members of this class are \onemax
\begin{align*}
    \OM(x) = \onemax(x) = \sum_{i=1}^n{x_i},
\end{align*}
which simply counts the number of bits set to \(1\) in \(x\), and \Binval
\begin{align*}
    \BinVal(x) = \sum_{i = 1}^n{2^i x_i},
\end{align*}
which returns the value of \(x\) read in base \(2\).

In this paper we focus on a few \emph{dynamic} monotone functions for which we show optimization is hard. When optimizing a dynamic monotone function \((f^t)_{t \ge 0}\), the evolutionary algorithm performs all fitness-comparisons at step \(t\) using \(f^t\), i.e.\ at each step we use a \emph{fresh} monotone function, so that \emph{locally}, optimization happens as for a classical monotone function. 
More concretely, for each $t \in \mathbb{N}$, we let $f^t:\{0,1\}^n \to \mathbb{R}$ be a monotone function that \emph{may depend on} $x^t$, and perform the selection step of Algorithm~\ref{alg:opoea} (or of any other optimization algorithm) with respect to $f^t$.

Here, we consider two dynamic adaptations of \Binval which we call \textsc{Advsersarial Dynamic BinVal} and \textsc{Friendly Dynamic BinVal} and which we respectively abbreviate to \adbv and \fdbv. Given some reference point \(x^t \in \{0, 1\}^n\), those are respectively defined as 
\begin{align*}
    \adbv_{x^t}(x) = \sum_{i=1}^{n}2^i x_{\sigma_A(i)}
    \qquad \text{and} \qquad 
    \fdbv_{x^t}(x) = \sum_{i=1}^{n}2^i x_{\sigma_F(i)},
\end{align*}
where \(\sigma_A = \sigma_{A,x^t}\) and \(\sigma_F=\sigma_{F,x^t}\) are permutations of \([1,n]\) that depend on $x^t$ and that satisfy\footnote{Note that there are several ways of choosing \(\sigma_A, \sigma_F\) satisfying those properties, as we may permute any \(0\)-bit with another \(0\)-bit, and similarly for the \(1\)-bits. As properties we study hold regardless of the precise selection of \(\sigma_A, \sigma_F\), we make a slight abuse of notation and talk of \emph{the} function \adbv and \emph{the} function \fdbv.} \(\sigma_A(i) \ge \sigma_A(j)\) whenever \(x^t_i \le x^t_j\) and \(\sigma_F(i) \le \sigma_F(j)\) whenever \(x^t_i \le x^t_j\). In other words, \textsc{Adversarial \dynBV} orders the bits in such a way that the binary value of \(x^t\) is \emph{minimized}, while \textsc{Friendly \dynBV} orders them so that the value of \(x^t\) is \emph{maximized}. Note that those definitions ensure that (and this is the property we are interested in), in the selection phase of the \opoea, every non-dominated offspring is accepted when optimizing \adbv; on the contrary, when optimizing \fdbv, we only accept offspring which dominates the parent.

The name \textsc{Adversarial \dynBV} was coined by Kaufmann, Larcher, Lengler and Zou~\cite{kaufmann2023self_TCS, kaufmann2022self_PPSN} because this choice of dynamic function makes it \emph{deceptively} easy to find a fitness improvement: as each \(1\)-bit of \(x^t\) has a lower weight than all \(0\)-bits, a mutation that improves the fitness as soon as a single \(0\)-bit flips to \(1\), meaning that there is a decently high chance of moving away from the optimum even though the fitness increases. This is for instance not the case with \onemax, where two \(1\)-bit flips outweigh a single \(0\)-bit flip and for which a step that increases the fitness necessarily reduces the distance to the optimum. Here we choose to name the second function \textsc{Friendly \dynBV} since the permutation is selected to satisfy the opposite property of \adbv. 

Our results are focused on a last dynamic monotone function which combines both functions above. More precisely, we define \textsc{Switching Dynamic BinVal}, abbreviated to \textsc{SDBV}, as follows 
\begin{align*}
    \sdbv_{x^t}(x) = 
    \begin{cases}
        \adbv_{x^t}(x) & \text{if } \OM(x) > n/2; \\
        \fdbv_{x^t}(x) & \text{if } \OM(x) \le n/2. 
    \end{cases}
\end{align*}
In this paper, whenever we run any of the evolutionary algorithms introduced in the previous section on \adbv, \fdbv or \sdbv, we always choose the reference point for those functions as the parent \(x^t\) at time \(t\). For this reason, we sometimes omit the subscript \(x_t\) and simply write \(\adbv(x), \fdbv(x), \sdbv(x)\) when the parent is clear from context.

\subsection{Further Tools}
In various parts of the proofs, we also use classical bounds or results from probability theory.
First, to switch between differences and exponentials, we will frequently make use of the following estimates, taken from Lemma~1.4.2 -- Lemma~1.4.8 in~\cite{doerr2020probabilistic}. 
\begin{lemma}
\label{lem:basic} \ 
    \begin{enumerate}[(i)]
        \item For all $r\ge 1$ and $0\le s\le r$,\label{lem:itm:basic-1}
        \begin{align*}
            (1-1/r)^{r} \le 1/e \le (1-1/r)^{r-1} 
            \quad \text{ and } \quad   
            (1-s/r)^{r} \le e^{-s} \le (1-s/r)^{r-s}.
        \end{align*}
        \item For all $0\le x \le 1$, it holds that $1-e^{-x} \ge x/2.$ \label{lem:itm:basic-2}
        \item For all $0 \le x\le 1$ and all $y\ge1$, it holds that $\tfrac{xy}{1+xy} \le 1- (1-x)^y \le xy. $ \label{lem:itm:basic-3}
\end{enumerate}
\end{lemma}

In our analysis of the runtime of the \opoea on \sdbv, we use the law of total variance. 
\begin{theorem}[Law of Total Variance]
\label{theo:LawOfTotalVariance}
    Let $X$ be a random variable with $\Var(X) < \infty$. Let $A_1, ..., A_n$ be a partition of the whole outcome space. Then
    \begin{align*}
        \operatorname{Var}(X) &= \sum_{i=1}^{n} \operatorname{Var}(X|A_i)\Pr(A_i) + \sum_{i=1}^{n} \EE[X|A_i]^2(1 - \Pr(A_i))\Pr(A_i) \\ 
        &\quad - 2\sum_{i=2}^{n}\sum_{j=1}^{i-1} \EE[X|A_i]\Pr(A_i) \EE[X|A_j]\Pr(A_j).
    \end{align*}
\end{theorem}
This runtime analysis also requires a version of Kolmogorov's maximal inequality for martingales.\footnote{A sequence $(X^t)_{t\ge 0}$ of random variables is a martingale if $\EE[|X^t|] <\infty$ and $\EE[X^t \mid X^1,\ldots,X^{t-1}] = \EE[X^t]$ for all $t \ge 0$. In a sub-martingale the latter condition is replaced by $\EE[X^t \mid X^1,\ldots,X^{t-1}] \ge \EE[X^t]$.} The theorem can be seen as a generalization of Chebychev's inequality with a built-in union bound and can be derived by observing that the square of a martingale is a sub-martingale and by then applying Doob's maximal inequality. 
\begin{theorem}[Kolmogorov's Maximal Inequality for Martingales~\cite{durrett2010probability}]
\label{theo:KolmogorovMartingale}
    Let $S^T = \sum_{t=1}^{T}X^t$  be a martingale with $\mathrm{E}[X^t] = 0$ and $\mathrm{Var}(X^t) < \infty$. Then for any $\beta > 0$,
    \begin{displaymath}
        \Pr\left[\underset{1 \le t \le T}{\max}  \lvert S^t \rvert \ge \beta \right] \le \frac{\mathrm{Var}(S^T)}{\beta^2}.
    \end{displaymath}
\end{theorem}

Finally, we use standard the following classical inequality due to Chernoff; the form we present is taken from Doerr~\cite{doerr2020probabilistic}.
\begin{theorem}[Chernoff Inequality]
    \label{thm:chernoff}
    \label{theo:ChernoffUpper}\label{theo:ChernoffLower}
    Let $X_1, \ldots, X_n$ be independent random variables taking values in $[0, 1]$, and define $X = \sum_{i=1}^{n} X_i$ as their sum. 
    \begin{enumerate}[(i)]
        \item For all \(\delta \in [0,1]\), \(\Pr[X \le (1 - \delta)\mathrm{E}[X]] \leq \exp\left(-\frac{\delta^2 \mathrm{E}[X]}{2}\right)\). \label{thm:itm:chernoff-lower}
        \item For all \(\delta \ge 0\), \(\Pr[X \geq (1 + \delta)\mathrm{E}[X]] \leq \exp{\left(-\frac{\delta^2 \mathrm{E}[X]}{2 + \delta}\right)}\). \label{thm:itm:chernoff-upper}
    \end{enumerate}
\end{theorem}

\section{\sdbv minimizes the Drift Towards the Optimum}
\label{sec:sdbv-minimizes-drift}

The goal of this section is to prove Theorem~\ref{thm:sdbv-minimizes-drift}. More precisely, we show that when optimizing with the \opoea, then of all \emph{dynamic} monotone functions, \sdbv is the one for which the drift is minimal at all points. 

We start the proof with a first lemma below; it states an equivalent condition which, albeit more technical, allows the rest of our proof to unfold.

\begin{lemma}
    \label{lem:sdbv-minimizes-drift-equivalent-condition}
    Let \(x \in \{0,1\}^n\) be an arbitrary search point. For \(i,j \ge 0\) define \(A_{i,j}\) as the set of all points that may be reached from \(x\) by flipping exactly \(i\) zero-bits to one, and \(j\) one-bits to zero.
    
    Assume that \(m\) is a monotone function such that for every monotone \(f\), and every \(i > j > 0\) the following quantity
    \begin{align}
        \sum_{y \in A_{i,j}} \left(\indicator{f(y) \ge f(x)} - \indicator{m(y) \ge m(x)} \right)
        - \sum_{y \in A_{j,i}} \left(\indicator{f(y) \ge f(x)} - \indicator{m(y) \ge m(x)} \right), \label{eq:sdbv-minimizes-drift-equivalent-condition} 
    \end{align}
    is non-negative, where $\indicator{\cdot}$ denotes the indicator function. Then \(m\) minimizes the drift at \(x\) for any mutation rate \(p\).
\end{lemma}

\begin{proof}
    Given an arbitrary search point \(x \in \{0,1\}^n\) and an arbitrary monotone function~\(f\), let \(Y\) be obtained from \(x\) by flipping each bit indepen\-dently with probability \(1/n\). The drift of \(f\) at \(x\) may be written as 
    \[
        \Delta_f(x) := \EE\left[ (Z(x) - Z(Y)) \cdot \indicator{f(Y) \ge f(x)} \right], 
    \]
    and to prove the lemma, it suffices to show that \(\Delta_f(x) - \Delta_{m}(x) \ge 0\) for all monotone \(f\). Using the partition of the space into \((A_{i,j})_{i,j\ge0}\), we may rewrite the drift of \(f\) as 
    \begin{align*}
        \Delta_f(x)
            & = \sum_{ i, j \ge 0 } \sum_{y \in A_{i,j}} (Z(x) - Z(y)) \cdot \Pr[Y = y] \cdot \indicator{f(y) \ge f(x)} \\
            &= \sum_{ i, j \ge 0} \sum_{y \in A_{i,j}} (i - j) \cdot p^{i+j} (1 - p)^{n-(i+j)} \cdot \indicator{f(y) \ge f(x)}.
    \end{align*}
    The contribution of \(A_{i,j}\) to the drift is \(0\) whenever \(i = j\), and by grouping together the terms \(A_{i,j}\) and \(A_{j,i}\) we obtain 
    \begin{align*}
        \Delta_f(x)
            &= \sum_{i > j \ge 0} (i - j) p^{i+j} (1 - p)^{n-(i+j)} \left[ \sum_{y \in A_{i,j}} \indicator{f(y) \ge f(x)} - \sum_{y \in A_{j,i}} \indicator{f(y) \ge f(x)} \right].
    \end{align*}
    Writing \(\alpha_{i,j} = (i - j) p^{i+j} (1 - p)^{n-(i+j)}\) for brevity, we then see that 
    \begin{align*}
        \Delta_{f}(x) - \Delta_m(x) 
            &= \sum_{i>j\ge0} \alpha_{i,j} \biggl( \sum_{y \in A_{i,j}}\left( \indicator{f(y) \ge f(x)} - \indicator{m(y) \ge m(x)} \right)  \\
            & \hspace{30mm} - \sum_{y \in A_{j,i}} \left( \indicator{f(y) \ge f(x)} - \indicator{m(y) \ge m(x)} \right)\biggr).
    \end{align*}
    Since \(f,m\) are both assumed to be monotone, we have \(\indicator{f(y) \ge f(x)} = \indicator{m(y) \ge m(x)} = 1\) whenever \(y \in A_{i, 0}\) and \(\indicator{f(y) \ge f(x)} = \indicator{m(y) \ge m(x)} = 0\) whenever \(y \in A_{0, i}\). In particular, all summands with \(j = 0\) are zero, and since the other are non-negative by assumption, the lemma is proved.
\end{proof}

Having proved Lemma~\ref{lem:sdbv-minimizes-drift-equivalent-condition}, we see that Theorem~\ref{thm:sdbv-minimizes-drift}  follows if we can prove that for all \(x \in \{0,1\}^n\)~\eqref{eq:sdbv-minimizes-drift-equivalent-condition} is non-negative when replacing \(m\) by \(\sdbv_x\). To work our way towards this result, we use the following combinatorial lemma which relates the elements of \(A_{i,j}\) to those of \(A_{j,i}\).

\begin{lemma}
    \label{lem:relation-Aij-Aji}
    Let \(x \in \{0,1\}^n\) be an arbitrary search point, \(i > j > 0\) and \(A_{i,j}, A_{j,i}\) as in Lemma~\ref{lem:sdbv-minimizes-drift-equivalent-condition}. 
    \begin{enumerate}[(i)]
        \item If \(Z(x) \le n/2\) then for every \(A \subseteq A_{i,j}\), there exists \(B \subseteq A_{j, i}\) such that the following holds. For all \(b \in B\), there is \(a \in A\) such that \(a \ge b\) and \(|B| \ge |A|\).
        \item If \(Z(x) \ge n/2\) then for every \(B \subseteq A_{j,i}\), there exists \(A \subseteq A_{i, j}\) such that the following holds. For all \(a \in A\), there is \(b \in B\) such that \(a \ge b\), and \(|A| \ge |B|\).

    \end{enumerate}
    
\end{lemma}

\begin{proof}
    Before proving the statement, it is useful to observe that every \(a \in A_{i,j}\) dominates exactly \(\binom{i}{i-j}\binom{n-Z(x)-j}{i-j}\) elements from \(A_{j,i}\). To see why this holds, recall that every element of \(A_{i,j}\) is obtained from \(x\) by flipping \(i\) zero-bits, and \(j\) one-bits. Hence, for \(a \in A_{i,j}\) to dominate \(b \in A_{j,i}\), it must be that all zero-bit flips in \(b\) are also zero-bit flips of \(a\) --- given \(a\), there are \(\binom{i}{i-j}\) ways to choose which zero-bit flips also appear in \(b\) --- and all of the \(j\) one-bit flips in \(a\) must also be one-bit flips from \(b\) --- given \(a\), there are \(\binom{n - Z(x) - j}{i - j}\) ways of choosing the additional flips in \(b\). Similarly, every \(b \in A_{j,i}\) must be dominated by \(\binom{i}{i-j}\binom{Z(x) - j}{i - j}\) elements from~\(A_{i,j}\).
    
    With those considerations at hand, we may now prove the first statement. Let \(x \in \{0,1\}^n\) be arbitrarily chosen such that \(Z(x) \le n/2\), let \(A\) be an arbitrary subset of~\(A_{i,j}\), and define \(B\) as the subset of all \(b \in A_{j,i}\) which are dominated by some \(a \in A\). We show that \(|B| \ge |A|\) by double-counting the number \(X\) of pairs \((a,b) \in A \times B\) with \(a \ge b\). Since each element \(a \in A\) dominates exactly \(\binom{i}{i-j}\binom{n - Z(x) - j}{i-j}\) elements, we must have \(X = |A| \cdot \binom{i}{i-j}\binom{n - Z(x) - j}{i-j}\). As we know that each \(b \in B\) may only be dominated by at most \(\binom{i}{i-j}\binom{Z(x) - j}{i - j}\) elements of \(A_{i,j}\), we must also have \(X \le |B| \cdot \binom{i}{i-j}\binom{Z(x) - j}{i - j}\). Combining gives 
    \begin{align*}
        |A| \binom{i}{i-j}\binom{n - Z(x) - j}{i-j} \le |B| \binom{i}{i-j}\binom{Z(x) - j}{i - j},
    \end{align*}
    which implies \(|A| \le |B|\) since \(Z(x) \le n/2\). This concludes the proof of the first item; the second follows from the symmetry of the argument.
\end{proof}

We may now prove the main result of this section. 

\begin{proof}[Proof of Theorem~\ref{thm:sdbv-minimizes-drift}]
    Let \(x \in \{0,1\}^n\) be an arbitrary search point and \(f\) an arbitrary monotone function. If we can prove that~\eqref{eq:sdbv-minimizes-drift-equivalent-condition} is non-positive for \(m = \sdbv_x\), then the theorem is proved. We split the analysis depending on whether \(Z(x) \le n/2\) or not.

    \proofitem{Case \(Z(x) < n/2\)} In this case we have \(\sdbv = \adbv\) and, by definition, \emph{every} non-dominated outcome is accepted. In particular,~\eqref{eq:sdbv-minimizes-drift-equivalent-condition} becomes 
    \begin{align*}
         \sum_{y \in A_{i,j}}\left( \indicator{f(y) \ge f(x)} - 1 \right) - \sum_{y \in A_{j,i}}\left( \indicator{f(y) \ge f(x)} - 1 \right).
    \end{align*}
   
    Letting \(A\) be the subset of those \(y \in A_{i,j}\) such that \(\indicator{f(y) \ge f(x)} = 0\), we know from Lemma~\ref{lem:relation-Aij-Aji} that there exists \(B \subseteq A_{j,i}\) with \(|B| \ge |A|\) such that for every \(b \in B\), there is \(a \in A\) with \(b \le a\). Since \(f\) is monotone, we must have \(f(b) \leq f(a)\), which implies \(\indicator{f(b) \ge f(x)} = 0\) for all \(b \in B\). In particular, we obtain 
    \begin{align*}
        \Delta_f(x) - \Delta_\sdbv(x)
            &= \sum_{a \in A}(-1) - \sum_{y \in A_{j,i}}(\indicator{f(y) \ge f(x)} - 1) \\
            &\ge \sum_{a \in A}(-1) - \sum_{b \in B}(- 1)
            \; = \; |B| - |A| \; \ge \; 0.
    \end{align*}

    \proofitem{Case \(Z(x) \ge n/2\)} The proof is very similar the the previous case. In this regime, we have \(\sdbv = \fdbv\) and every non-dominating offspring is rejected. In particular, we have 
    \begin{align*}
         \Delta_f(x) - \Delta_\sdbv(x) = \sum_{y \in A_{i,j}} \indicator{f(y) \ge f(x)} - \sum_{y \in A_{j,i}} \indicator{f(y) \ge f(x)}.
    \end{align*}
    Now letting \(B\) be the set of those \(y \in A_{j,i}\) such that \(\indicator{f(y) \ge f(x)}=1\), by Lemma~\ref{lem:relation-Aij-Aji} there exists a set \(A \subseteq A_{i,j}\) dominating \(B\) and such that \(|A| \ge |B|\). By monotonicity of \(f\), we derive 
    \(\Delta_f - \Delta_\sdbv \ge |A| - |B| \ge 0\).
\end{proof}

\section{\sdbv does not Maximize Optimization Time}
\label{sec:proof-not-longest-hitting-time}

In the previous section we showed that the drift, i.e.\ the expected amount by which we move \emph{towards} the optimum, is minimized in all points by \sdbv (among the class of all dynamic monotone functions). For this reason, it is tempting to believe that \sdbv is the function for which the expected optimization time is extremal. 

However, this is not true in general. In this section, we show by direct computation for odd $n\in[9,45]$ that \sdbv is \emph{not} the function that maximizes the runtime. Runtime analysis is usually conducted using drift analysis, but as we have established previously that \sdbv minimizes the drift, this approach will not allow us to prove this result. 

To establish our result, we compute the expected hitting time of \sdbv and other related functions exactly for small values of $n$. The expected optimization time of a function \(f\) when started at some \(x \in \{0,1\}^n\) may be expressed as an affine combination of the expected runtimes when started at all other \(y \in \{0,1\}^n\). We employ computational methods to solve this system of equations, and are able to get (exact) numerical values for the expected running time when the dimension of the problem is small. We detail our approach further, but we first start with some intuition of why minimizing is not the same as maximizing the expected runtime.

\subsection{Intuition: Jumps Are Beneficial}
In Section \ref{sec:sdbv-minimizes-drift}, we showed that whenever $Z^t<n/2$, a drift-minimizing function leads to the acceptance of as many offspring as possible - while maintaining monotonicity at each step, whereas at times $t$ such that $Z^t>n/2$ this is achieved by the rejection of as many offspring as possible. At the point $Z^t= n/2$, both strategies are pessimal in the sense that they lead to the same - minimized - drift. Recall that we have opted for FDBV in this case.  Surprisingly, while the number of potential accepted offspring when $Z^t= n/2$ does not affect the drift, it does play a role for the expected optimization time. It turns out to be beneficial for the optimization of a function to accept more offspring. 

For an intuitive understanding, consider the following toy example. Assume that for some function $f_1$ we could only gain progress of one at $Z^t = n/2$ (i.e. $Z^t-Z^{t+1} = 1$), with some probability $p$, while otherwise, the state remains unchanged. For a function $f_2$, we also make progress of one with probability $p$, but we have the additional options $Z^t-Z^{t+1} = k$ and $Z^t-Z^{t+1} = -k$, both with some probability $q>0$. Then the drift is identical, but we want to argue that it is still easier to optimize $f_2$ than $f_1$. The reason is that progress gets harder as we approach the optimum - evidenced by a drift bound of $\Theta(s^2/n^2)$ proven below. Therefore it is harder to progress from $Z^t =n/2$ to $Z^t = n/2 -k$ than to progress from $Z^t =n/2+k$ to $Z^t = n/2$. Hence, if we jump $k$ towards the optimum, we gain more (namely, the expected time to proceed from $Z^t =n/2$ to $Z^t = n/2 -k$) than we lose if we jump $k$ away from the optimum (the expected time to proceed from $Z^t =n/2+k$ to $Z^t = n/2$). Therefore, optimizing $f_2$ takes less expected time than optimizing $f_1$. 

Returning to SDBV, this suggests that we can increase the expected optimization time by expanding the interval in which we use FDBV, thus accepting many offspring in a slightly larger range. This is indeed what we will do in the proof.

\subsubsection{Proof Details: A Simplified Version of SDBV}
In the characterization of SBDV employed so far, in principle every search point $x\in \{0,1\}^n$ induces a different function. This would require up to $ 2^n$ states to be modeled as a Markov chain and would lead to computational problems even for small problem sizes. We will show in the following, that it is enough to consider the number of zeros in the parent, which allows us to analyze an equivalent randomized process that requires only $n+1$ states. We begin by making the following observations, restricting to the case of ADBV by assuming that $Z^t<n/2$ - the case of FDBV is analogous:
\begin{itemize}
    \item One iteration of the loop can be described by first mutating the parent, followed by the selection step.
    \item By definition, ADBV is equivalent to some permutation $\sigma$ of the bit string evaluated the Binary Value function. Hence an equivalent step can be performed by applying the permutation, performing Binary Value selection and reversing the permutation, that is by applying $\sigma^{-1}$.
    \item Since bits are mutated independently, the expected optimization time remains unchanged if the bits are permuted before mutation. 
    \item Reversing the mutation and the permutation step guarantees that after the permutation, the bits of the offspring will be sorted non-decreasingly - informally, all zero-bits are sorted to the front of the string, followed by the one-bits.
    \item This shows that the permutation $\sigma$ is not unique: any permutation which sorts the bit string non-decreasingly works. 
    \item Now since between the application of the inverse transformation $\sigma^{-1}$ and the non-decreasing sorting of the mutated bit string in the next iteration no selection occurs, applying $\sigma^{-1}$ becomes obsolete.
    \item It follows that all the relevant information about the fitness of a search point is contained in the number of zero-bits it contains.
\end{itemize}
We can therefore analyze the equivalent Algorithm~\ref{algo:simplified}:\\

\begin{algorithm}
\caption{Optimization of SDBV }

\SetKwInput{Init}{Initialization}
\SetKwInput{Sort}{Sorting}
\SetKwInput{Opt}{Optimization}
\SetKwInput{GC}{Generational Change}
\SetKwInput{Mut}{Mutation}
\SetKwInput{Sel}{Selection}

\Init{Set $x^0\in \{0,1\}^n$ uniformly at random}
\Opt{
    \For{$t=0,1,\dots$}{
\Sort{
\If{$ZM(x^t)<n/2$} {Sort $x^t$ non-decreasingly}
\Else{Sort $x^t$ non-increasingly}
}

\GC{$y \leftarrow x^t$}
\Mut{Flip each bit in $y$ independently with probability $1/n$}
\Sel{
\If{$\BinVal(y)\ge \BinVal(x^t)$}{$x^{t+1} \leftarrow y$}
\Else{$x^{t+1} \leftarrow x^t$}
}
}

}
\end{algorithm}
\label{algo:simplified}

\subsubsection{Calculating Optimization Times} The optimization can now be modeled as a Markov chain with the state at time $t$ given by the number of zeros in $x^t$ and the expected optimization time the expected number of steps required to reach state $0$. Denote by $p_{a,b}$ the transition probability of reaching state $a$ from state $b$ in one step and as previously by $s$ the number of zeros in $x^t$. Let further $G$ and $R$ be the random variables which record the number of flips from $x^t$ to $y$ of zero-bits and of one-bits respectively. We refer to them as \emph{green flips} and \emph{red flips} respectively. 

First, let $s \ge n/2$ and let $k$ be the next state with $s \neq k$. Clearly, the state can only change if the offspring $y$ is accepted, which means by the definition of FDBV that there are no red flips in $y$. Thus, to get from state $s$ to state $k$ exactly $k-s$ green flips and $0$ red flips must occur in $y$. The probability of that event can be easily computed using the fact that $G$ and $R$ are independent and binomially distributed. Hence for $s \ge n/2$ and $k \neq s$
\begin{align*}
    p_{s,k} &= \Pr(G = k-s, R = 0 \mid Z^t = s) \\
    &= \Pr(G = k-s \mid Z^t = s) \cdot \Pr(R = 0 \mid Z^t = s) \\
    &=\binom{s}{s-k} \cdot \Big(\frac{1}{n}\Big)^{s-k} \cdot \Big(1-\frac{1}{n}\Big)^{k} \cdot \binom{n-s}{0} \cdot  \Big(\frac{1}{n}\Big)^{0} \cdot \Big(1-\frac{1}{n}\Big)^{n-s} \ \\
    &= \binom{s}{s-k} \cdot \Big(\frac{1}{n}\Big)^{s-k} \cdot \Big(1-\frac{1}{n}\Big)^{n-s+k}.
\end{align*}
Observe that for $k > s$, the probability is always $0$.

Now, let $s < n/2$ and let $k$ be the next state with $s \neq k$. For the offspring to be accepted, there must be at least one green flip in $y$. We compute
\begin{align*}
p_{s,k} &=  \Pr(Z^{t+1} = k \mid Z^t = s) \\
&= \Pr(s-Z^{t+1} = s-k \mid Z^t = s) \\
&= \Pr(G - R = s-k \mid Z^t = s, G \ge 1) \cdot \Pr(G \ge 1 \mid Z^t = s)\\
&= \sum_{i=1}^{s} \Pr(G = i \mid Z^t = s) \cdot \Pr(R = i-(s-k) \mid Z^t =s) \\
&= \sum_{i=1}^{s}{\binom{s}{i} \Big(\frac{1}{n}\Big)^i \Big(1-\frac{1}{n}\Big)^{s-i} \cdot  \binom{n-s}{i-(s-k)}  \Big(\frac{1}{n}\Big)^{i-(s-k)}  \Big(1-\frac{1}{n}\Big)^{n-s-(i-s+k)}} \\
&= \sum_{i=1}^{s}{\binom{s}{i} \cdot \binom{n-s}{i-s+k} \cdot \Big(\frac{1}{n}\Big)^{2i-s+k} \cdot \Big(1-\frac{1}{n}\Big)^{n-2i+s-k}}.
\end{align*}

Finally, the probability of remaining in a state $s$ is given by
\begin{displaymath}
    p_{s,s} = 1 - \sum_{k=0}^{s-1}{p_{s,k}} - \sum_{k=s+1}^{n}{p_{s,k}}.
\end{displaymath}

After calculating the transition probabilities, one can formulate recursive equations to compute the hitting times. Given a current state $s$, denote the expected time to reach state $0$ by $\mathrm{E}[T_s]$. Clearly $\mathrm{E}[T_0] = 0$. For $s > 0$ it holds that
\begin{displaymath}
    \mathrm{E}[T_s] = 1 + \sum_{k = 0}^{n}{p_{s,k} \cdot \mathrm{E}[T_s]} = 1 + \sum_{k = 1}^{n}{p_{s,k} \cdot \mathrm{E}[T_k]}.
\end{displaymath}

These $n$ recursive equations can now be combined into a single system of linear equations, which can be solved to extract $\mathrm{E}[T_s]$ for all $s > 0$. For example, when $n=3$ the corresponding matrix equation that need to be solved is
\[
\begin{bmatrix}
p_{1,1}-1 & p_{1,2} & p_{1,3}  \\
p_{2,1} & p_{2,2}-1 & p_{2,3}  \\
p_{3,1} & p_{3,2} & p_{3,3}-1 \\
\end{bmatrix}
 \cdot
\begin{pmatrix}
\mathrm{E}[T_1] \\
\mathrm{E}[T_2] \\
\mathrm{E}[T_3] \\
\end{pmatrix}
=
\begin{pmatrix}
-1 \\
-1 \\
-1 \\
\end{pmatrix}.
\]
Finally, one can compute $\mathrm{E}[T]$, the total expected optimization time by using law of total expectation with the conditioning on the number of zeros in the initial search point. We compute
\begin{align*}
\mathrm{E}[T] &= \sum_{k = 0}^{n}{\mathrm{E}[T \mid \ZM(x^0) = k] \cdot P[\ZM(x^0) = k]} \\ &= \sum_{k = 0}^{n}{\mathrm{E}[T_k] \cdot P[\ZM(x^0) = k]} \\
&= \sum_{k=0}^{n}{\mathrm{E}[T_k] \cdot \binom{n}{k} \cdot \frac{1}{2^n}}.
\end{align*}

We proceed to show that the hardest dynamic monotone function to optimize on the $(1+1)$-EA is not necessarily the one with lowest drift. We have established that whenever $\ZM(x^t) \leq n/2$, \textsc{Adversarial \dynBV}, which assigns the weights of the Binary Value function to \emph{minimize} the fitness of $x^t$, minimizes the drift. Conversely, when $\ZM(x^t) \geq n/2$, \textsc{Friendly \dynBV}, which assigns the weights of the Binary Value function to \emph{maximize} the fitness of $x^t$, attains the lowest drift. \textsc{Switching \dynBV} alternates between ADBV and FDBV based on the number of zeros in $x^t$: it follows the behavior of ADBV when $\ZM(x^t) < n/2$ and that of FDBV otherwise. We will demonstrate that a slight reduction in this cutoff --- which results in an increase in drift --- can further increase the optimization time. This is because for $\ZM(x^t) \approx n/2$, both ADBV and FDBV result in similar drift, but their behavior during optimization is fundamentally opposite: While ADBV accepts as many offspring as possible for a monotone function, FDBV rejects as many offspring as possible for a monotone function. 
To demonstrate the effect of modifying this cutoff, we compute the precise hitting times for fixed problem sizes. Thus, all calculations have to be done using fractions without any rounding. 
The code for this can be found at \href{https://github.com/OliverSieberling/SDBV-EA}{https://github.com/OliverSieberling/SDBV-EA}. For readability, the final results are then rounded to 4 digits. \par
We will now look at $n = 9$, as it is the smallest $n$ where the desired effect is observable. By the proof in Section \ref{sec:sdbv-minimizes-drift}, SDBV is drift-minimizing, which selects by ADBV when $\ZM(x^t) < 4.5$ and selects by FDBV when $\ZM(x^t) \ge 4.5$. We compare this with functions with a smaller cutoff for switching from ADBV to FDBV. Let $\bigtriangleup_s$ denote the drift when the optimization is in state $s$, that is, when $\ZM(x^t)=s$. Then, reducing the cutoff must increase the drift at the modified states by Theorem~\ref{thm:sdbv-minimizes-drift}. We verify this theoretical result for $n=9$ by computing the drifts $\bigtriangleup_s$ numerically in Table~\ref{tab:driftn=9}. 
\begin{table}[h]
  \centering
  \caption{Drift for $n=9$}
  \label{tab:driftn=9}
  \begin{tabular}{c|c|c|c}
    \textbf{} & \textbf{SDBV - cutoff 4.5} & \textbf{SDBV - cutoff 3.5} &\textbf{SDBV - cutoff 2.5} \\
    \hline
    $\bigtriangleup_0$ & $0$ & $0$ & $0$ \\
    $\bigtriangleup_1$ & $0.01235$ &  $0.01235$ & $0.01235$\\
    $\bigtriangleup_2$ & $0.05898$ & $0.05898$ & $0.05898$\\
    $\bigtriangleup_3$ & $0.13489$ & $0.13489$ & $\textbf{0.16442}$ \\
    $\bigtriangleup_4$ & $0.23572$ & $\textbf{0.24664}$ & $\textbf{0.24664}$\\
    $\bigtriangleup_5$ & $0.34683$ & $0.34683$ & $0.34683$ \\ 
    $\bigtriangleup_6$ & $0.46822$ & $0.46822$  & $0.46822$ \\
    $\bigtriangleup_7$ & $0.61454$ &  $0.61454$ & $0.61454$\\
    $\bigtriangleup_8$ & $0.79012$ & $0.79012$ & $0.79012$ \\
    $\bigtriangleup_9$ & $1$ & $1$ & $1$
  \end{tabular}
\end{table}

Intuitively, this increase in drift could be expected to be beneficial for faster optimization. However, for a cutoff of $3.5$ the expected hitting time does not decrease; it increases, see Table~\ref{tab:example}. This effect vanishes for an even lower cutoff of~$2.5$. 

\begin{table}[h]
\centering
\caption{Hitting times for $n=9$}
\label{tab:example}
  \begin{tabular}{c|c|c|c}
    \textbf{} & \textbf{SDBV - cutoff 4.5} & \textbf{SDBV - cutoff 3.5} &\textbf{SDBV - cutoff 2.5} \\
    \hline
    $\mathrm{E}[T_0]$ & $0$ & $0$ & $0$ \\
    $\mathrm{E}[T_1]$ & $30.1845$ & $30.1861$ & $30.0440$\\
    $\mathrm{E}[T_2]$ & $41.2612$ & $41.2646$ & $40.9707$ \\
    $\mathrm{E}[T_3]$ & $47.1214$ & $47.1276$ & $46.3839$ \\
    $\mathrm{E}[T_4]$ & $50.7524$ & $50.7716$ & $50.1061$\\
    $\mathrm{E}[T_5]$ & $53.3796$ & $53.3959$ & $52.7251$ \\ 
    $\mathrm{E}[T_6]$ & $55.3045$ & $55.3210$ & $54.6501$\\
    $\mathrm{E}[T_7]$ & $56.7601$ & $56.7766$ & $56.1057$\\
    $\mathrm{E}[T_8]$ & $57.8835$ & $57.9000$ & $57.2291$ \\
    $\mathrm{E}[T_9]$ & $58.7644$ & $58.7809$ & $58.1100$\\
    \hline
    $\mathrm{E}[T]$ & $50.9855$ & $50.9997$ & $50.3553$ 
  \end{tabular}
\end{table}

We now investigate empirically how the optimal cutoff changes with increasing $n$.
We observe that the parity of $n$ plays a significant role. If $n$ is even, there is a middle state $n/2$, where both ADBV and FDBV selection are drift-minimizing. If $n$ is odd for both states $\lceil{x/2} \rceil$ and $\lfloor {x/2} \rfloor$ the respective drift-minimizing strategy is unique. 

\begin{figure}[H]
\begin{minipage}[c]{0.46\linewidth}
     \centering
  \includegraphics[width=\textwidth]{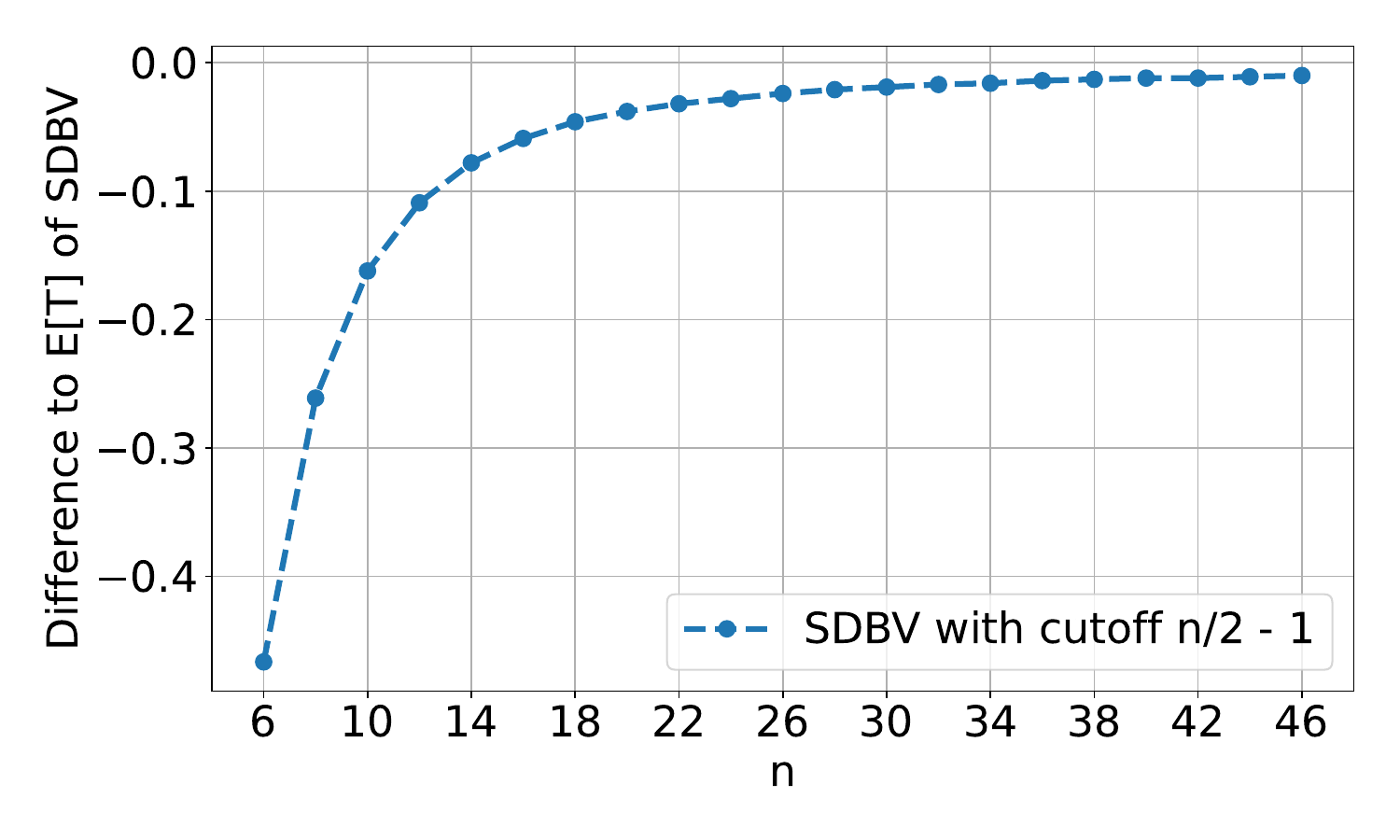}
\end{minipage} \hfill
 \begin{minipage}[c]{0.46\linewidth}
     \centering
  \includegraphics[width=\textwidth]{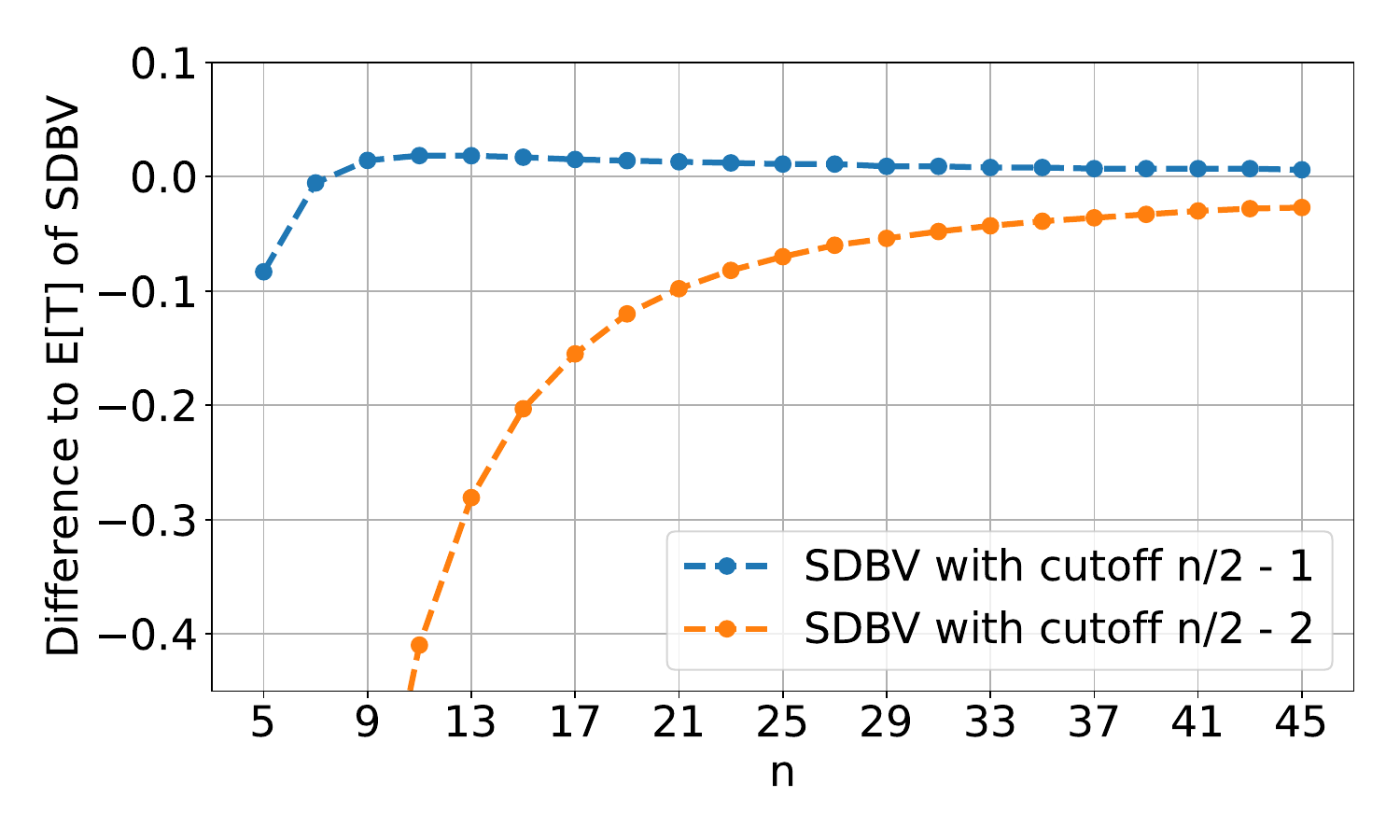}
\end{minipage} \hfill
 
  \caption{Change in expected optimization time by reducing the cutoff for even (left) and odd (right) $n$}
  \label{fig:example}
\end{figure}

First, consider even $n$. Figure~\ref{fig:example} shows that for small $n$, even the slightest decrease in the cutoff does not increase the hardness.
On the other hand, for all tested odd $n \ge 9$, decreasing the cutoff by $1$ to $n/2-1$ increases the expected hitting time, see Figure~\ref{fig:example}. This effect disappears for cutoff $n/2-2$. 

Modifying the cutoff by some fixed amount changes the behavior only for a fixed amount of states. With increasing $n$, these modified states become less relevant for the entire optimization. Therefore, it is not surprising that the expected hitting time of SDBV with decreased cutoff approaches the expected hitting time of SDBV as $n \rightarrow \infty$. 

\begin{figure}[H]
  \centering
  \includegraphics[width=0.6\textwidth]{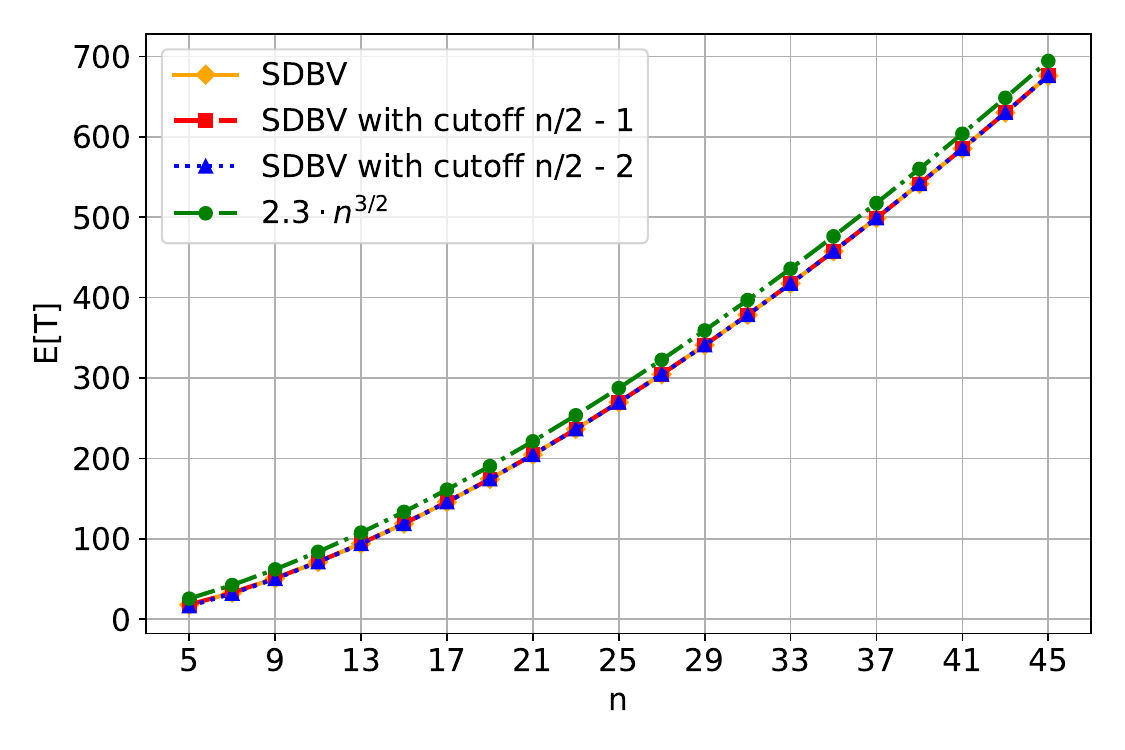}
  \caption{Expected optimization times for SDBV with different cutoffs}
  \label{fig:example3}
\end{figure}

Finally, in Figure~\ref{fig:example3} we look at the absolute expected hitting times. This serves two purposes. First, to display how minor the above effect is. Second, to illustrate the $\Theta(n^{3/2})$ runtime of the $(1+1)$-EA on SDBV, which we prove formally in Section~\ref{sec:profn32}.

\section{The Expected Optimization Time of \sdbv is \(\Theta(n^{3/2})\)}\label{sec:profn32}

In the previous two sections we have established first that \sdbv minimizes the drift towards the optimum, and then showed that this does not imply that it maximizes the runtime. However, in this section we will prove that the runtime on SDBV is $\Theta(n^{3/2})$, thus establishing Theorem~\ref{thm:sdbv-runtime}. The upper bound for the runtime follows from the general upper bound of $O(n^{3/2})$ in the \poea framework.\footnote{Actually, the \poeam; see~\cite{colin2014monotonic} for the details.} Since this upper bound holds for \emph{all} dynamic montone functions, it also proves that SDBV gives \emph{asymptotically} the highest runtime. 
To obtain these results, it thus suffices to prove the following lemma.

\begin{lemma}
    \label{lem:sdbv-runtime-LB}
    When started on a uniformly random point \(x^0 \in \{0,1\}^n\), the \opoea requires \(\Omega(n^{3/2})\) steps in expectation to optimize \sdbv. 
\end{lemma}

The rest of this section is dedicated to proving Lemma~\ref{lem:sdbv-runtime-LB}. Since the statement is asymptotic, whenever it is needed, we may assume that \(n\) is sufficiently large. Our proof follows that of the lower bound on the runtime of the \poea by Colin, Doerr and F\'erey~\cite{colin2014monotonic}. While their paper only contains a sketch of the proof, we choose to provide a detailed proof for completeness. We start with a lemma which estimates the drift of \sdbv at every point. While an upper bound on this drift would be sufficient to derive Lemma~\ref{lem:sdbv-runtime-LB}, we also provide a lower bound estimate. As we have established that \sdbv minimizes the drift among dynamic monotone functions, this lower bound holds not only for \sdbv, but for any such function, which we believe to be interesting on its own. 

\begin{lemma}
    \label{lem:sdbv-drift-estimate}
    Consider the \opoea on \sdbv. Let  \(x^t\) be the parent at time \(t\), and \(\Delta^t = \EE\left[ Z(x^t) - Z(x^{t+1}) \mid x^t\right]\) the drift at time \(t\). Then for every \(t \ge 0\), we have  \(\Delta^t = \Theta\left( (Z(x^t) / n)^2 \right)\).
\end{lemma}

\begin{proof}
    Consider a given \(x^t \in \{0,1\}^n\), and let \(y\) be obtained from \(x^t\) by flipping each bit independently with probability \(1/n\). For brevity, we write \(z^t, z^{t+1}\) for \(Z(x^t), Z(x^{t+1})\). \sdbv has two regimes depending on whether \(z^t \le n/2\) or not, and we deal with both cases separately. 

    \proofitem{Case \(z^t \ge n/2\)} 
    In this regime \(z^t\) and \(n\) are of comparable magnitude so it suffices to prove that $\Delta^t = \Theta(1)$. Since $\Delta^t$ is upper bounded by the expected number of bit flips, we have $\Delta^t \le 1$. For the lower bound, recall that when \(z^t \le n/2\) we have \(\sdbv = \fdbv\), so \(y\) is rejected unless it dominates \(x^t\). Hence, the distance from the optimum cannot increase, $z^{t+1} \le z^t$. Now consider the event that exactly one zero-bit is flipped, but no one-bit. This has probability $z^t\cdot 1/n \cdot (1-1/n)^{n-1} = \Omega(1)$, and hence $\Delta^t = \Omega(1)$.
    
    \proofitem{Case \(z^t < n/2\)}
    In this regime we have \(\sdbv = \adbv\), meaning that the offspring \(y\) is accepted if and only if it is non-dominated by \(x^t\), i.e.\ if and only if at least one \(0\)-bit is flipped. Let us call \(\cE_i\) the event in which \(i\) zero-bits are flipped: one easily checks that \(\Pr[\cE_i] = \binom{z}{i} n^{-i} (1 - 1/n)^{z^t - i} = \Theta\left( \binom{z}{i} n^{-i} \right)\), since \((1 - 1/n)^{z^t - i} \in [e^{-1}, 1]\). Independent of \(\cE_i\), the number of one-bits flipped back to \(0\) is \((n - z^t) / n = 1 - z^t/n\), which implies \(\EE[z^{t+1} \mid \cE^i] = z^t - i + (1 - z^t/n)\). Summing up, we obtain that 
    \begin{align*}
        \Delta^t 
            \; &= \; \sum_{i=1}^{z^t} \Pr[\cE_i]  \left( z^t - \EE[z^{t+1} \mid \cE_i ]\right)
            \; = \; \Theta\left( \sum_{i=1}^{z^t} \binom{z^t}{i} n^{-i} (i - 1 + z^t/n) \right).
    \end{align*}
    The first term of this sum is of order \(\Theta((z^t/n)^2)\), so the lower bound on the drift is proved. To prove the upper bound, we observe that for every \(i \ge 2\), the \(i\)-th term in the sum is upper bounded by \((z^t/n)^{i} / (i-1)! \le (z^t / n)^2 / (i-1)!\), and since \(\sum_{i \ge 2} \frac{1}{(i-1)!}\) converges, the whole drift must be bounded by \(O((z^t / n)^2)\). 
\end{proof}

In the rest of the argument, we show that traversing a region around \(Z(x^t) \approx \sqrt{n}\) already takes an expected \(\Omega(n^{3/2})\) steps. For the reasoning to work, we first need to prove at some time during the optimization the search point satisfies this, i.e.\ that we neither start with fewer than \(\sqrt{n}\) or `jump' above this interval. This is captured by the following lemma.

\begin{lemma}
    \label{lem:HighProbFallsSqrtn}
    With high probability, the optimization takes \(n^{3/2}\) steps, or there exists some step of the optimization at which the number of zero-bits of the parent falls in the interval $[\sqrt{n}, 2\sqrt{n}]$.
\end{lemma}

\begin{proof}
    Note that the event holds if we can show that 1) the starting point is chosen with at least \(2 \sqrt{n}\) zero-bits and 2) in none of the first \(n^{3/2}\) steps the number of zeros jumps by more than \(\sqrt{n}\).
    
    By Chernoff's Inequality (Theorem~\ref{thm:chernoff}) the probability that the first event does not hold is at most
    \begin{flalign*}
        \Pr\left[\Bin(n, 1/2) < \sqrt{n} \right] = e^{-\Omega(n)} = o(1).
    \end{flalign*}
    The number of zero-bits flipped at each step is dominated by \(\Bin(n, 1/n)\). In particular, by Chernoff's Inequality the probability that this happens at some step is at most 
    \begin{align*}
        \Pr\left[ \Bin(n, 1/n) \ge \sqrt{n} \right] = e^{-\Omega(\sqrt{n})}.
    \end{align*}
    Even after union bound over the first \(n^{3/2}\) steps the probability remains super-polynomially small. Finally, another use of union bound shows that both events happen simultaneously with probability \(1 - o(1)\).
\end{proof}

We have just established that, with high probability, at some point during the optimization we have a search point \(x^t \in \{0,1\}^n\) with a number of zero-bits contained in \([\sqrt{n}, 2\sqrt{n}]\), and we denote by \(t_0\) the first time this happens. Using the drift computed previously, and Kolmogorov's Maximal Inequality (Theorem~\ref{theo:KolmogorovMartingale}), we now wish to show that reaching a new state \(x^{t_1}\) which has either no zero-bits, or more than \(3\sqrt{n}\) zero-bits takes an expected \(\Omega(n^{3/2})\) steps. Let $T$ denote the number of such additional steps. If we are able to show this, then Lemma~\ref{lem:sdbv-runtime-LB} follows.

To be able to apply Kolmogorov's inequality and prove this result, we first need to restate our problem as a martingale problem. To this intent, we define \(X^t := Z(x^{t_0+t}) - Z(x^{t_0+t+1})\) as the variation of the number of zero-bits \(t\) steps after \(t_0\), as well as \(Y^t := X^t - \EE[X^t \mid X^1, \dots, X^{t-1}]\). With those definitions, we easily check that \(\sum_{t} Y^t\) is a martingale. In the following lemma, we compute the variance of each term.

\begin{lemma}
    \label{lem:variance-y}
    Let \(t < T\), then \(\Var [Y^t \mid Y^1, \dots, Y^{t-1}] = O(1/\sqrt{n})\).
\end{lemma}

\begin{proof}
    Conditioned on \(X^0, \dots, X^{t-1}\) (or equivalently on \(Y^0, \dots, Y^{t-1}\)), \(X^t\) and \(Y^t\) differ by an additive constant \(\EE[X^t \mid X^0, \dots, X^{t-1}]\), so they have the same variance. Also, the variance of \(X^t\) conditioned on \(X^0, \dots, X^{t-1}\) is clearly upper bounded by \(\EE[ (X^t)^2 \mid X^0, \dots, X^{t-1}]\), so it suffices to show that this quantity is bounded by \(O(n^{-1/2})\).

    Define \(\cE^t\) as the event that at least one \(0\)-bit is flipped at step \(t\). As we assume that the number of \(0\)-bits is smaller than \(3 \sqrt{n} \le n/2\), \sdbv behaves like \adbv and the child generated at this step is accepted if and only if \(\cE^t\) happens. In particular, \(X^t = 0\) whenever \(\overline{\cE^t}\) happens. We may then use the Law of Total Variance (Theorem~\ref{theo:LawOfTotalVariance}) to express
    \begin{align*}
        \Var[X^t]  &= \Var [X^t \mid \cE^t] \cdot \Pr[\cE^t] + \EE[X^t \mid \cE^t]^2 \cdot \Pr[\overline{\cE^t}] \cdot \Pr[\cE^t],
    \end{align*}
    where we omitted the conditioning on \(X^0, \dots, X^{t-1}\) for brevity. As the number of zero-bits is upper bounded by \(3 \sqrt{n}\), we have 
    \[
        \Pr[\cE^t] \le 1 - (1-1/n)^{3\sqrt{n}} = O(1/\sqrt{n}),
    \]
    by Lemma~\ref{lem:basic}~\ref{lem:itm:basic-3}. Since the total number of bits is binomially distributed with parameter \(1/n\) (i.e.\ the tails of the distribution decay exponentially fast), we have \(\Var [X^t \mid \cE^t] = O(1)\) and \(\EE[X^t \mid \cE^t]^2 = O(1)\), which concludes the proof.
\end{proof}

We now have all the ingredients and may prove Lemma~\ref{lem:sdbv-runtime-LB}.

\begin{proof}[Proof of Lemma~\ref{lem:sdbv-runtime-LB}]
    By Lemma~\ref{lem:HighProbFallsSqrtn}, we know that with high probability there exists some time at which the number of zeros of the parent is in the range \([\sqrt{n}, 2\sqrt{n}]\). Let \(t_0\) be the first time when this happens, \(T\) the additional number of steps required, and let \(X^t, Y^t\) be defined as previously.

    As we start with a number of zero-bits in the interval \([\sqrt{n}, 2\sqrt{n}]\) and since \(X^t\) represents the variation of the number of zeros at time \(t\), we must have
    \begin{align}
        \left| \sum_{t=0}^{T} X^t \right| \ge \sqrt{n}. \label{eq:sdbv-runtime-condition}
    \end{align}
    To prove the lemma, we show that this inequality has constant probability of not being satisfied within the first \(\Omega(n^{3/2})\) steps after \(t_0\), which implies that \(T\) (and therefore the total expected optimization time) has expectation at least \(\Omega(n^{3/2})\). From Lemma~\ref{lem:sdbv-drift-estimate} and Lemma~\ref{lem:variance-y} we know that \(\EE[X^t \mid X^0, \dots, X^{t-1}] \le c_1 / n\) and \(\Var[Y^t \mid X^0, \dots, X^{t-1}] \le c_2 / \sqrt{n}\) for absolute constants \(c_1, c_2\). Let \(\delta = \min \left\{1, 1/2c_1, 1/8c_2 \right\}\) and \(\bar t = \delta n^{3/2}\). During all steps \(t \in [0, \bar t]\), the random variables $X^t$ and $Y^t$ differ by at most $c_1/n$, and hence \(\sum_{t=0}^{\tau}{X^t}\) and \(\sum_{t=0}^{\tau} Y^t\) differ by at most \(\bar t \cdot c_1 / n \le \tfrac{\sqrt{n}}{2}\) for all $\tau \in [0,\bar t]$. In particular, for~\eqref{eq:sdbv-runtime-condition} to happen before step \(\bar t\), it must be that \(\left|\sum_{t =0}^\tau Y^t\right| \ge \sqrt{n} / 2\) happens for some \(\tau < \bar t\) steps. Using Kolmogorov's inequality, this probability is at most 
    \begin{align*}
        \Pr[T \le \bar t] \; \le \; \Pr\left[ \max_{\tau \le \bar t} \left| \sum_{t=1}^{\tau} Y^t \right| \; \ge \; \sqrt{n} / 2 \right]
            &\le 4 \cdot \frac{\Var \left[ \sum_{t=1}^{\bar t} Y^t \right]}{n}.
    \end{align*}
    A simple induction gives \(\Var \left[ \sum_{t=1}^{\bar t} Y^t \right] \le \bar t \cdot c_2 / \sqrt{n} \le n / 8\), from which we deduce \(\Pr[ T \le \bar t] \le 1/2\). This immediately implies \(\EE[T] \ge \bar t / 2\), which finishes the proof.
\end{proof}

\section{Simulations}\label{sec:simulations}

This section aims to provide empirical support to our theoretical results.\footnote{The code for the simulations can be found at \href{https://github.com/OliverSieberling/SDBV-EA}{https://github.com/OliverSieberling/SDBV-EA}.
} In particular, we verify our theoretical results for the \opoea and test them on the static \oplea and \oclea with \(\lambda=2 \ln n\), and the \saolea. The selection of algorithms is motivated as follows: The  \((1+\lambda)\)-EA is the natural extension of the  \((1+1)\)-EA, where in each generation \(\lambda\) offspring are produced. This allows parallelization. Its non-elitist cousin \((1,\lambda)\) has been shown to escape local optima more efficiently~\cite{hevia2021self,jorritsma2023comma}. Finally the \saolea allows to dynamically adjust the offspring population size at runtime to the difficulty of the current optimization phase. In this setting, Kaufmann, Larcher, Lengler and Zou also formulated the conjecture that ADBV was the hardest dynamic monotone function to optimize. 
 For each algorithm, we investigate problem sizes from \(n=20\) to \(n=420\) in increments of \(20\). For the \saolea, the update
strength is set to \(F = 1.15\) and the success ratio to \(s=0.25\). Across all algorithms, the mutation rate is set to \(1/n\). We always start with a uniformly random bitstring. The algorithm terminates when the optimum \(1^n\) is found. All plots are averaged over $500$ runs.

\begin{figure}[H]
    \begin{minipage}[c]{0.46\linewidth}
        \centering
         \includegraphics[width=\textwidth]{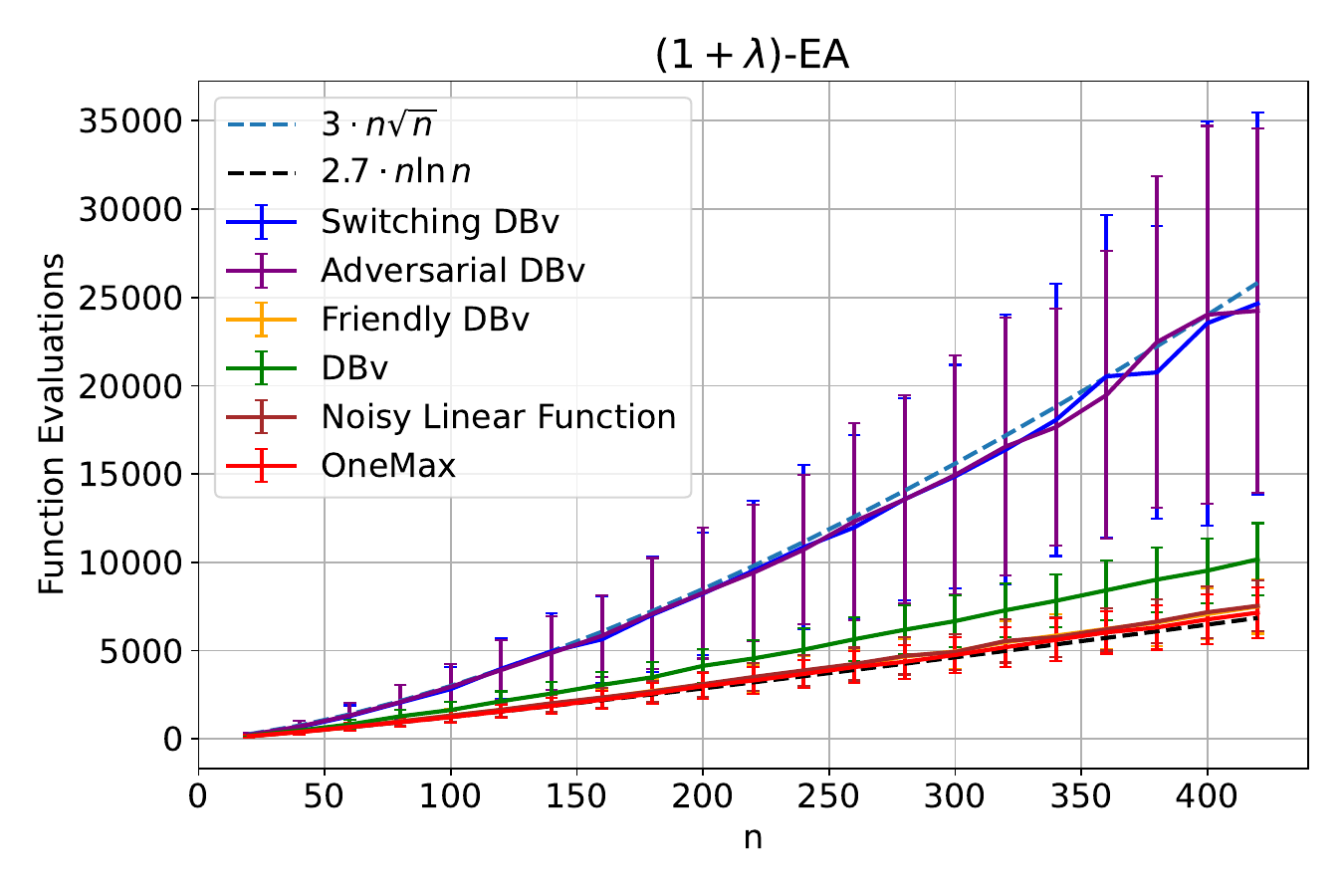}
    \end{minipage} \hfill
     \begin{minipage}[c]{0.46\linewidth}
        \centering
         \includegraphics[width=\textwidth]{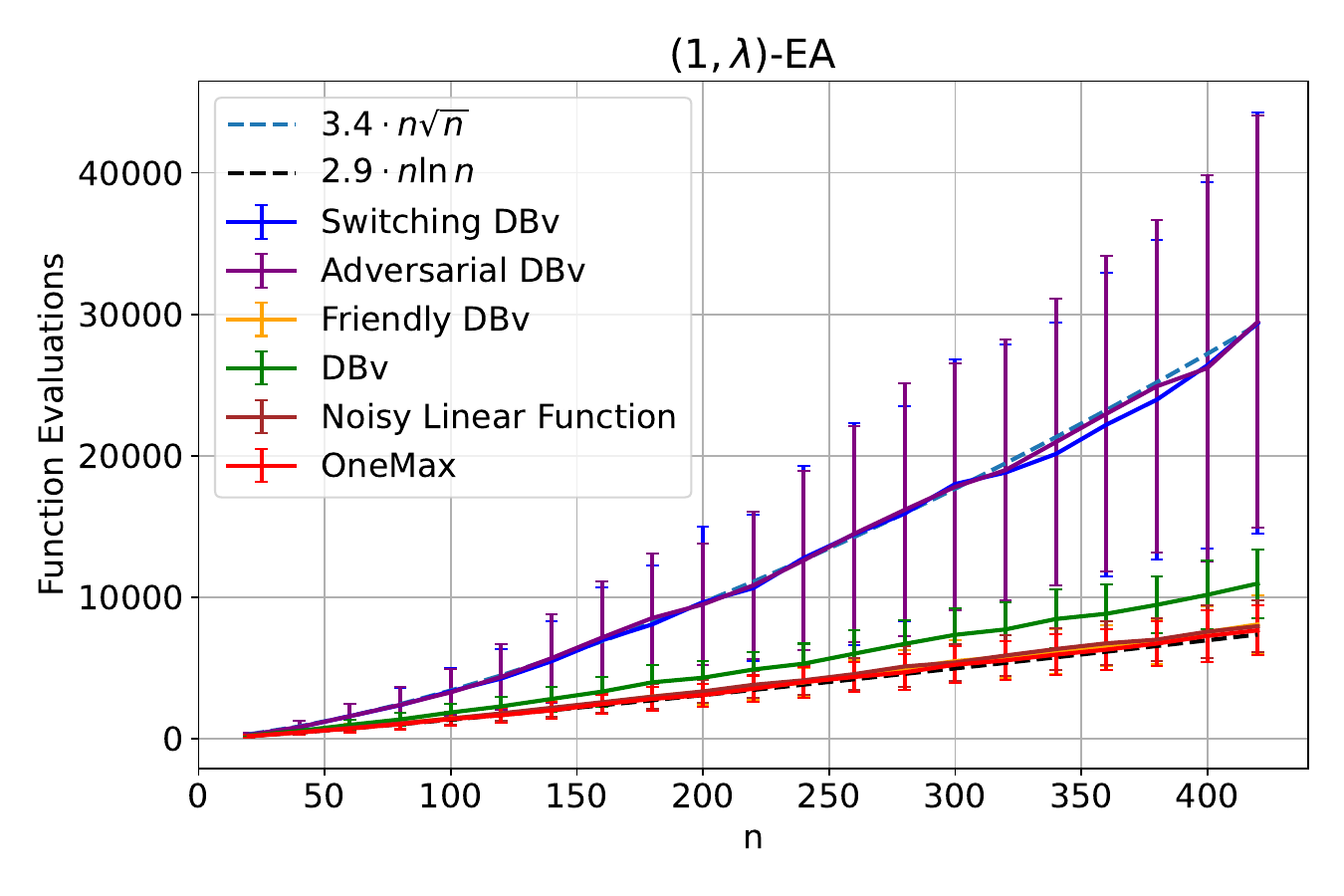}
    \end{minipage} 
    \begin{minipage}[c]{0.46\linewidth}
        \centering
         \includegraphics[width=\textwidth]{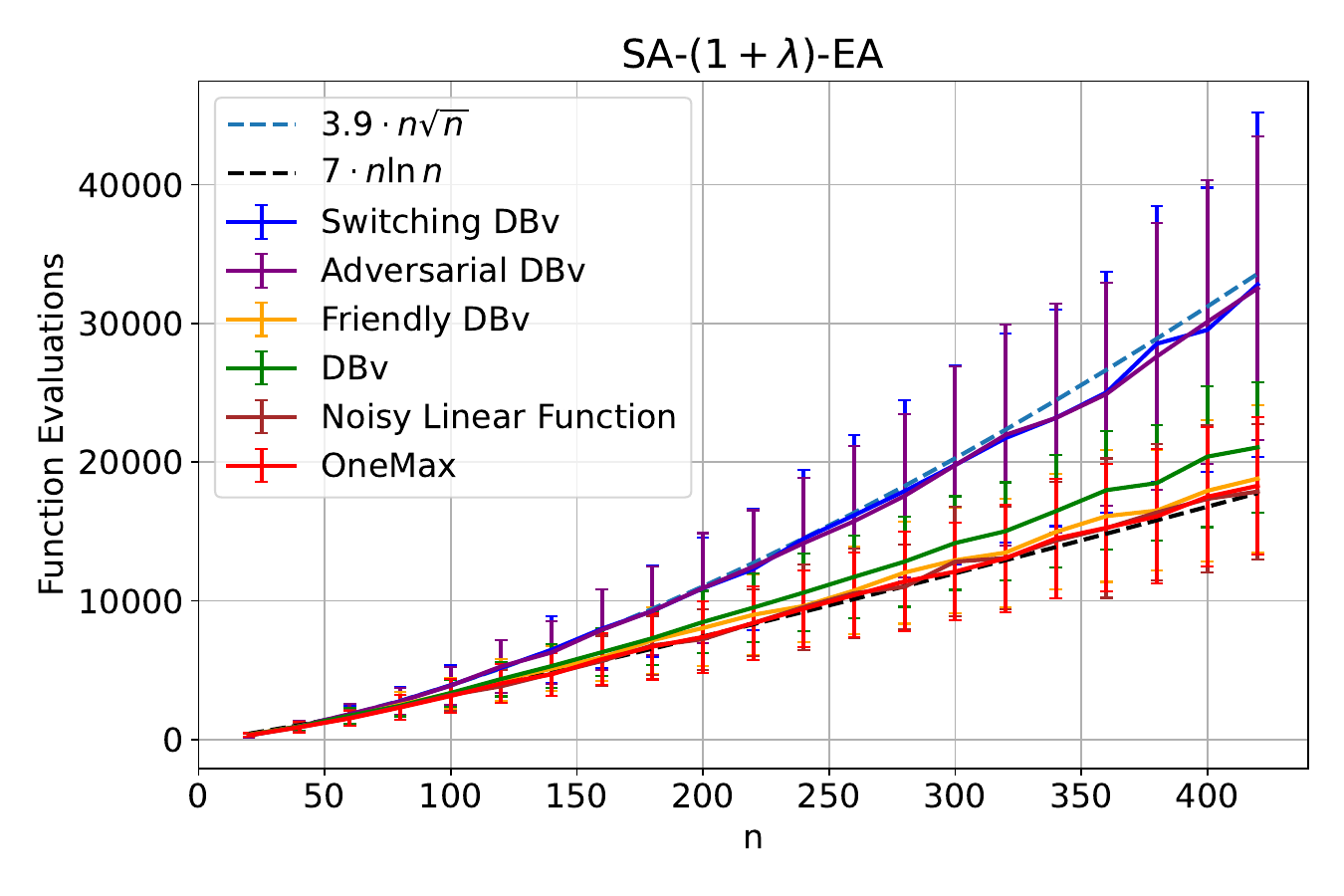}
    \end{minipage} \hfill
     \begin{minipage}[c]{0.46\linewidth}
        \centering
         \includegraphics[width=\textwidth]{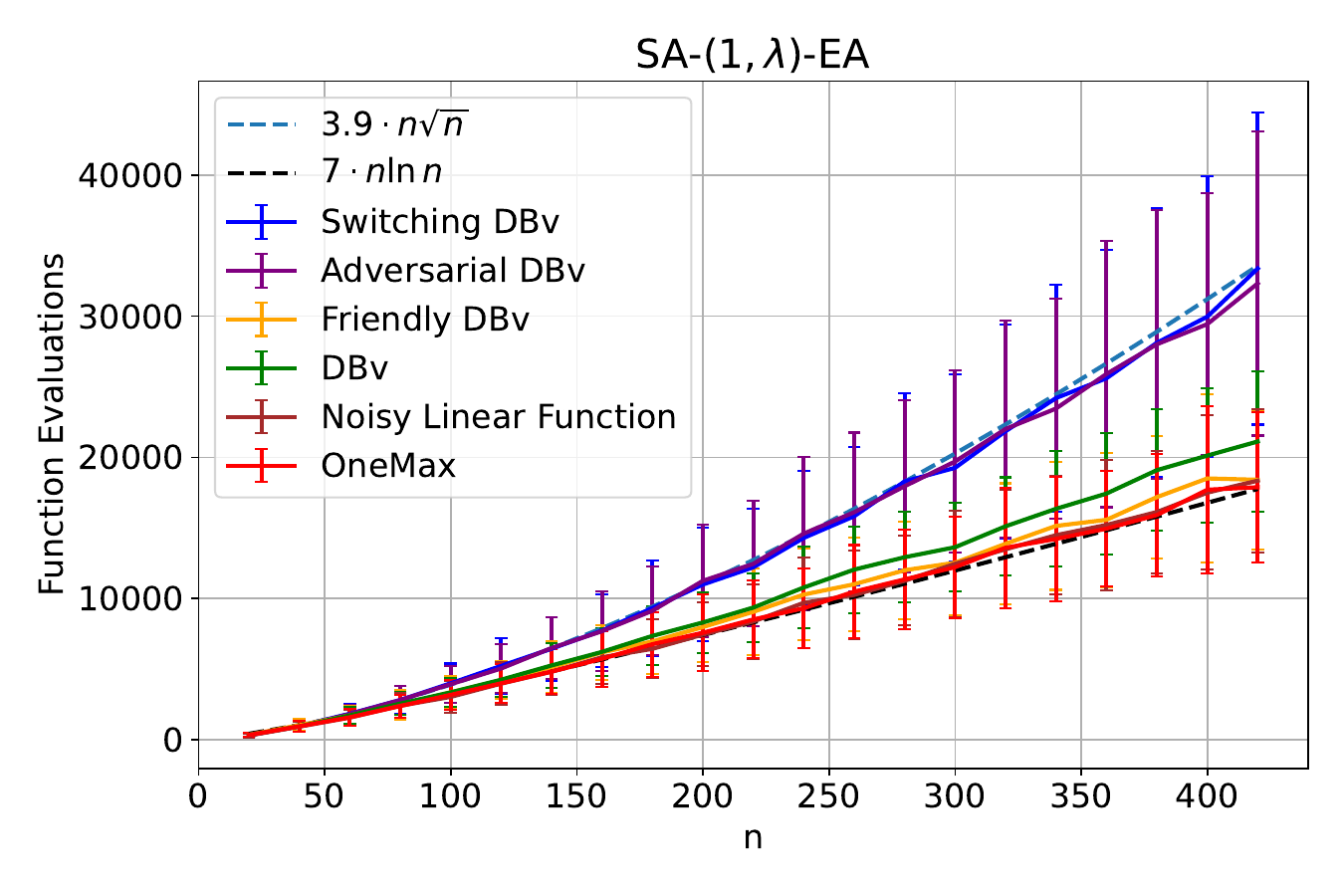}
    \end{minipage} 
    \caption{Runtime comparisons for the $(1+\lambda)$-EA, the $(1,\lambda)$-EA and their self adjusting variants. Depicted are the average number of function evaluations over 500 runs and the standard deviations.
\label{plot:lambda}
}
\end{figure}

As our benchmarks, we evaluate the following: the building blocks of SDBV, that is ADBV and FDBV, the PO-EA, which for the \ooea is asymptotically equivalent to SDBV, and the \poeam. We further include DBV and noisy linear functions (with discrete uniform distribution over $[n]$) which are two well-researched dynamic monotone functions, as well as the PO-EA. We refer the interested reader for definitions of these functions and background to \cite{kaufmann2023self_TCS}, \cite{lengler2018noisy} and \cite{colin2014monotonic} respectively. Finally, we include OneMax, the “drosophila” of functions in the study of evolutionary algorithms.
\begin{figure}[h]
    \begin{minipage}[c]{\linewidth}
        \centering
         \includegraphics[width=0.6\textwidth]{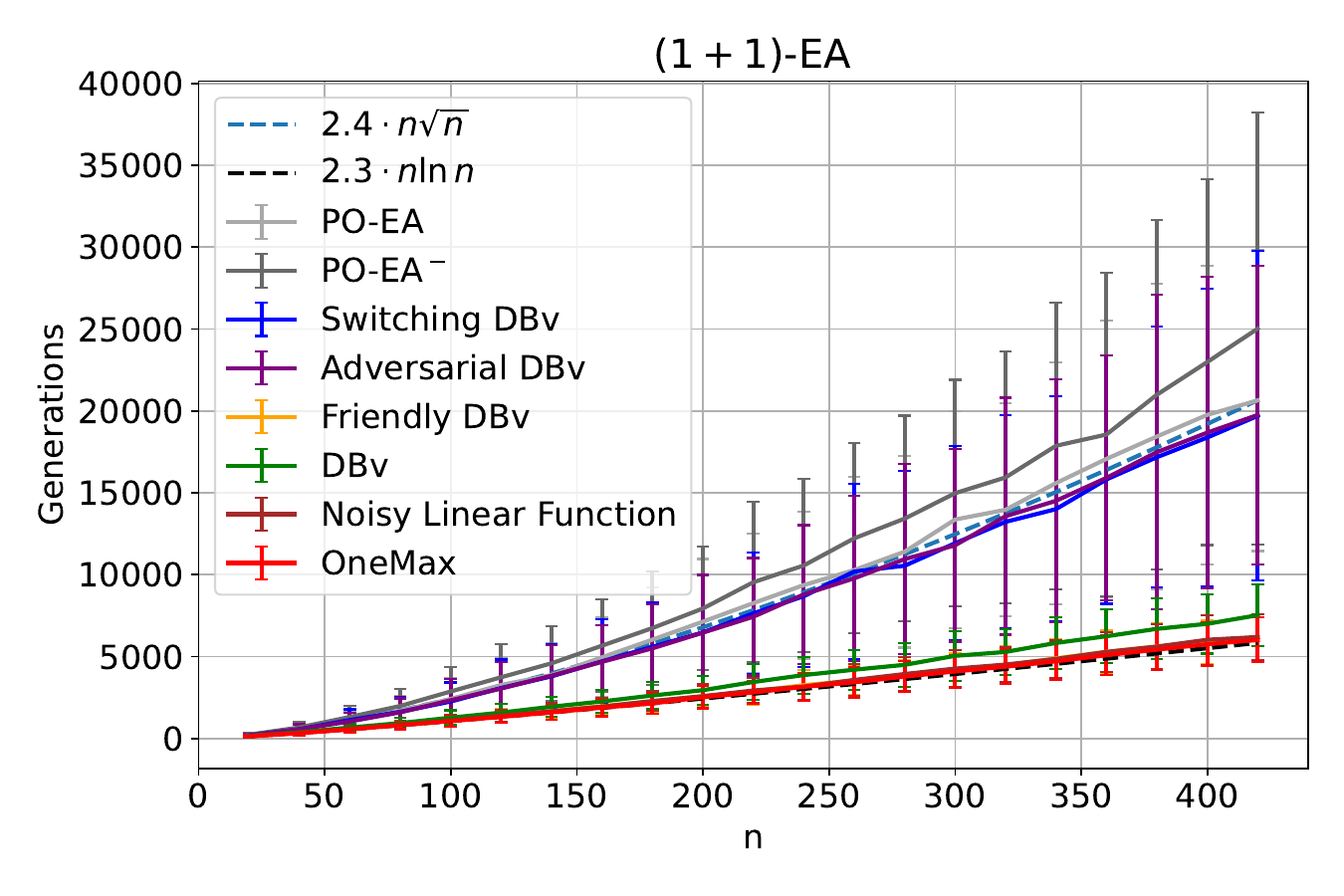}
         \caption{Runtime comparison for the $(1+1)$-EA. Depicted is the average over 500 runs and the standard deviation (not to be confused with confidence intervals, which are much smaller).\label{plot:oneplusone}
}
    \end{minipage} \hfill
\end{figure}
As we see in Figures~\ref{plot:oneplusone} and~\ref{plot:lambda}, SDBV and ADBV track each other closely. It is also visible in particular in the algorithms with multiple offspring - both elitist and non-elitist, that asymptotically there is little difference between the SDBV and the ADBV. That is, differences in the function far away from the optimum play little role asymptotically, even when the selection behavior is the polar opposite, already for moderate problem sizes. This behavior is evident across algorithms, giving credence to the conjecture that SDBV realizes the pessimism of the PO-EA for a large class of algorithms. Figure~\ref{plot:oneplusone} shows that up to a slightly smaller multiplicative constant, SDBV indeed matches the runtime even of the fully pessimistic variant of the PO-EA.  Finally, Figue~\ref{plot:lambda} shows that the \(\Theta(n^{3/2})\) runtime also holds for the much larger class of algorithms beyond the $(1+1)$-EA, making this the first known dynamic monotone function with runtime \(\omega(n \log n)\) in a setting where these algorithms are known to be efficient. In particular, in our experiments we also see a separation between the runtime of dynamic monotone functions such as for Dynamic Binary Value and Noisy Linear where randomization occurs naively and whose runtime is known to lie in $O(n \log n)$ generations for the regime where optimization can be proven to be efficient \cite{lengler2018noisy,kaufmann2023self_TCS}. That is, dynamic monotone functions with "uniform" randomization appear intrinsically easier than SDBV.\\

\section{Conclusion}\label{sec:conclusion}

We have shown that at least in the context of dynamic monotone functions, the PO-EA framework is not overly pessimistic but that there exist indeed functions which pose the difficulties ascribed abstractly to an evolutionary algorithm. We have proved this for the case of the classical \((1+1)\)-EA and our experiments show that this holds for general offspring sizes, hence also for the \((1+\lambda)\)-EA as well as for some comma-strategies, both the static \((1,\lambda)\)-EA and the self-adjusting \((1,\lambda)\)-EA require the same asymptotic optimization time of \(\Theta(n^{3/2})\) generations. Our function SDBV which materializes this pessimism also minimizes drift. It is however, when conducting a precise runtime analysis for small dimensions, not the function with highest expected optimization time. Which function indeed maximizes expected optimization time is still open. As we can see from our experiments, the difference in runtime to the drift-minimizing function SDBV decreases for increased dimension. This motivates our concluding conjecture.

\begin{conjecture}
    The expected number of generations required by the \ooea to optimize the hardest dynamic monotone functions exceeds that for SDBV by at most $o(n)$ generations.
\end{conjecture}

\bibliography{references}

\end{document}